\documentclass{article}

\usepackage{natbib}%
\usepackage[figuresright]{rotating}

\usepackage{url}
\usepackage{amsmath,amsthm}

\theoremstyle{definition}

\theoremstyle{remark}

\usepackage[ruled, vlined]{algorithm2e}
\usepackage{graphicx}
\usepackage{diagbox}
\usepackage{xcolor}
\usepackage{colortbl}
\usepackage{pgfgantt}
\usepackage{subfig}

\definecolor{j0}{rgb}{0.43,0.67,0.28}
\definecolor{j1}{rgb}{0.92,0.49,0.19}
\definecolor{j2}{rgb}{1,0.75,0}
\definecolor{j3}{rgb}{0.65,0.65,0.65}
\definecolor{j4}{rgb}{0.4,0.58,0.75}

\newcommand{\RowColor}{\rowcolor{gray!50} \cellcolor{white}}
\newcommand{\ie}{\textit{i.e., }}
\newcommand{\IGl}{IG$_{\mbox{{\scriptsize SJ}}}$}
\newcommand{\IIGl}{IIG$_{\mbox{\scriptsize SJ}}$}
\newcommand{\LS}{Local-Search}
\newcommand{\Cmax}{$C_{\mbox{max}}$}

\usepackage[french,textsize=small]{todonotes}

\begin{document}

% Title
\title{Exploiting Promising Sub-Sequences of Jobs \\ to solve the No-Wait Flowshop Scheduling Problem}

%Authors, affiliations address.
\author{Lucien Mousin, Marie-El\'eonore Kessaci and Clarisse Dhaenens\\Univ. Lille, CNRS, Centrale Lille, UMR 9189 - CRIStAL - \\ Centre de Recherche en Informatique Signal et Automatique de Lille,\\ F-59000 Lille, France
\\ lucien.mousin@univ-lille.fr
\\ marie-eleonore.kessaci@univ-lille.fr
\\ clarisse.dhaenens-flipo@univ-lille.fr
}

%\date{December 05, 2018}

%\affil{\affmark{a}Univ. Lille, CNRS, Centrale Lille, UMR 9189 - CRIStAL - \\ Centre de Recherche en Informatique Signal et Automatique de Lille, F-59000 Lille, France}
%\email{lucien.mousin@ed.univ-lille1.fr; me.kessaci@univ-lille1.fr; clarisse.dhaenens@univ-lille1.fr}

%Abstract

%Keywords, etc.
%\keywords{heuristic; learning; No-wait Flowshop}

\maketitle

\begin{abstract}
The no-wait flowshop scheduling problem is a variant of the classical permutation flowshop problem, with the additional constraint that jobs have to be processed by the successive machines without waiting time. 
To efficiently address this  NP-hard combinatorial optimization problem we conduct an analysis of the structure of good quality solutions. 
This analysis shows that the No-Wait specificity gives them a common structure: they share identical sub-sequences of jobs, we call \textit{super-jobs}. 
After a discussion on the way to identify these super-jobs, we propose \IGl, an algorithm that exploits super-jobs within the \emph{state-of-the-art} algorithm for the classical permutation flowshop, the well-known Iterated Greedy (IG) algorithm. 
An iterative approach of \IGl~ is also proposed.
Experiments are conducted on Taillard's instances.
The experimental results show that exploiting super-jobs is successful since
\IGl~ is able to find out 64 new best solutions. 
\end{abstract}

\section{Introduction}
%%% General
%Scheduling problems are classical combinatorial optimization problems. Most of them are NP-hard, and many advanced optimization methods (exact, heuristics and meta-heuristics) are proposed to deal with them.
%Within these approaches, some of them integrate some learning phases to guide the search to good quality regions of the search space. 
Scheduling problems represent an important class of combinatorial optimization problems, most of them being \emph{NP-hard}. 
They consist in the allocation of different operations on a set of machines over the time. 
The aim of scheduling problems is to find the schedules (or solutions) that optimize one or more criteria such as the makespan or the flowtime. 

Among these scheduling problems, the jobshop and the flowshop have been widely studied in the literature, as they represent many industrial situations. 
In this paper, we are specifically interested in the No-Wait Flowshop Scheduling Problem (NWFSP)~\citep{Rock_1984}. 
This extension of the classical Permutation Flowshop Scheduling Problem (PFSP) imposes that operations have to be processed without any interruption between consecutive machines. 
This constraint describes real process constraints that may be found in the chemical industry for example. 
It incorporates a specific structure on solutions we will exploit during the solving. 
Therefore we will conduct an analysis on some instances, of good quality solutions that will allow us to make an observation: sub-sequences of jobs are common within these solutions. 
This will lead us to define the concept of \emph{super-jobs} (sub-sequences of consecutive jobs) and to exploit them into a \LS~method~\citep{Hoos_2004}.

Indeed, while solving \emph{NP-hard} problems, the use of exact methods, mostly based on enumerations, is not practicable. Therefore heuristics and metaheuristics are developed.
Heuristics designed for a specific problem can use, for example, priority rules to  construct the schedule. Their advantages are their speed -- most of them are greedy heuristics -- and their specificity -- characteristics of the problem to be solved can be exploited --. 
Metaheuristics, on their side, are generic methods that can be applied to many optimization problems. Their efficiency is linked to their high ability to explore the search space and the way the metaheuristic has been adapted to the problem under study. In this context, the aim of the present work is to study how the identification of super-jobs increases the performance of a metaheuristic, and more specifically a \LS, up to obtain new best-known solutions.

 %The initial solution, when one is required, may also influence the quality of the final solution provided.

%The approach we propose here can enter within this category. Indeed, in this paper we are interested in solving the No-Wait Flowshop Scheduling Problem (NWFSP) which is a variant of the well-known Permutation Flowshop Scheduling Problem (PFSP). 

%%% Motivations
Indeed, as it will be exposed in the article, the concept of \emph{super-jobs} has several advantages. 
\begin{itemize}
	\item First, it reduces the combinatorics of the problem -- by reducing the search space -- and makes feasible the use of efficient methods for small size problems.
	\item Second, it modifies the fitness landscape during the execution of the \LS~and thus, avoid some local optima.
\end{itemize}
Experiments show that exploiting these super-jobs obtains very good results as a large number of new best solutions have been found out for large size instances of the commonly used Taillard's benchmark~\citep{Taillard_1993}.\\

%% Plan
To expose the way we propose to define and exploit super-jobs, this article is organized as follows: Section~\ref{sec:nwfsp} presents the No-Wait Flowshop Scheduling Problem and a brief literature review on the problem. Section~\ref{sec:superjob} reports the analysis we conducted and the super-jobs are formally defined. Section~\ref{sec:method} describes the proposed approach to exploit super-jobs, and Section~\ref{sec:experiments} gives experimental results. Then, Section~\ref{sec:iteratedsuperjob} improves the method with an iterated version and shows its performance. The last section presents some conclusions as well as some perspectives.

%%%%%%%%%%%%%%%%%%%%%%%%%%%%%%%%%%%%%%%%
%%%%%%%%%%%%%%%%%%%%%%%%%%%%%%%%%%%%%%%%
\section{No-Wait Flowshop Scheduling Problem}

Among the numerous scheduling problems, the Flowshop Scheduling Problem (FSP) consists in scheduling a set of \(N\) jobs \(\{J_1,\dots,J_N\}\), 
on a set of \(M\) machines \(\{M_1,\dots,M_M\}\). Several versions exist according to specific constraints that may be considered. In this article we are interested in the No-Wait version of this problem.

\subsection{Problem Description}
\label{sec:nwfsp}
%%%%%%%%%%%%%%%%%%%%%%%%%%%%%%%%%%%%%%%%
%%%%%%%%%%%%%%%%%%%%%%%%%%%%%%%%%%%%%%%%

\paragraph{Presentation of the problem} The No-Wait Flowshop Scheduling Problem (NWFSP) is a variant of the well-known Permutation Flowshop Scheduling Problem (PFSP), where no waiting time is allowed between the processing of a job on the successive machines~\citep{Rock_1984}. 
Despite this constraint, the NWFSP remains a NP-hard problem.

More formally, the NWFSP may be defined as follows.
Let $J$  be a set of $N$ jobs that have to be processed on a set of $M$ ordered machines;
%$p_{i,j}$ is the \textit{processing time} of a job $i$ on a machine $j$. 
Machines are critical resources that can only process one job at a time. 
A job $J_i$ is composed of $M$ tasks \(\{t_{i,1},\dots,t_{i,M}\}\) for the $M$ machines respectively. A processing time \(p_{i,j}\) is associated to each task \(t_{i,j}\).
As for the PFSP, the sequence of jobs is the same on each machine,  hence, a solution of the NWFSP is commonly represented by  a permutation $\pi=\{\pi_1, \dots, \pi_N\}$ where $\pi_1$ is the first job scheduled and $\pi_N$ the last one.
In this paper, the goal is to find a sequence that minimizes the makespan (\(C_{\mbox{max}}\)) defined as the total completion time of the schedule (equation~(\ref{eq_ob})). 
\begin{equation}
  \label{eq_ob}
  f(\pi) = C_{\mbox{max}} = \max_{i \in \{1,\dots,N\}}\{C_{i}\} = C_{\pi_N} 
\end{equation}

\paragraph{No-Wait variant specificity} The NWFSP possesses a characteristic not present in the classical PFSP that enables to reduce the computation time of the makespan of a sequence. 
This characteristic concerns the delay defined between two consecutive jobs as the start-up interval between the two jobs of a sequence on the first machine.
Let $d_{i,i'}$ be the delay between two jobs $i$ and $i'$; it is computed as follows:
\begin{equation} 
	d_{i,i'} = p_{i,1} + \max_{1 \le r \le M} \left( \sum^r_{j=2} p_{i,j} - \sum^{r-1}_{j=1} p_{i',j}, 0 \right)
\end{equation}

 \begin{figure}
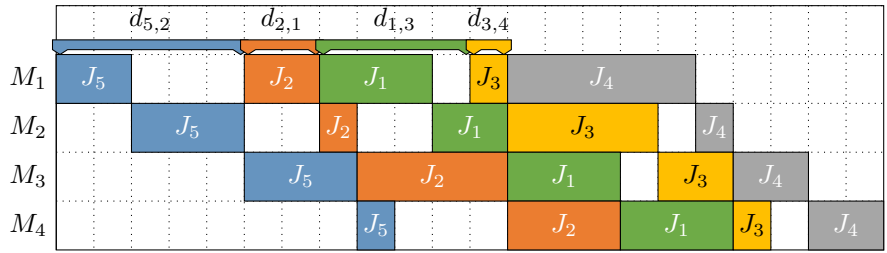

 \centerline{
\begin{ganttchart}[
vgrid,
hgrid,
group top shift=0.7,
y unit chart = 0.65cm,
bar height=1,
bar top shift=0]{0}{21}
%tasks
\ganttgroup[group/.append style={draw=black, fill=j4}, inline]{$d_{5,2}$}{0}{4}
\ganttgroup[group/.append style={draw=black, fill=j1}, inline]{$d_{2,1}$}{5}{6}
\ganttgroup[group/.append style={draw=black, fill=j0}, inline]{$d_{1,3}$}{7}{10}
\ganttgroup[group/.append style={draw=black, fill=j2}, inline]{$d_{3,4}$}{11}{11}\\
\ganttbar{$M_1$}{0}{0} 
\ganttbar[bar/.append style={fill=j4}, bar label font=\color{white} , inline]{$J_5$}{0}{1} 
\ganttbar[bar/.append style={fill=j1}, bar label font=\color{white} , inline]{$J_2$}{5}{6}
\ganttbar[bar/.append style={fill=j0}, bar label font=\color{white} , inline]{$J_1$}{7}{9}
\ganttbar[bar/.append style={fill=j2}, bar label font=\color{black} , inline]{$J_3$}{11}{11} 
\ganttbar[bar/.append style={fill=j3}, bar label font=\color{white} , inline]{$J_4$}{12}{16} \\
\ganttbar{$M_2$}{0}{-1}
\ganttbar[bar/.append style={fill=j4}, bar label font=\color{white} , inline]{$J_5$}{2}{4} 
\ganttbar[bar/.append style={fill=j1}, bar label font=\color{white} , inline]{$J_2$}{7}{7}
\ganttbar[bar/.append style={fill=j0}, bar label font=\color{white} , inline]{$J_1$}{10}{11} 
\ganttbar[bar/.append style={fill=j2}, bar label font=\color{black} , inline]{$J_3$}{12}{15}
\ganttbar[bar/.append style={fill=j3}, bar label font=\color{white} , inline]{$J_4$}{17}{17} \\
\ganttbar{$M_3$}{0}{-1}
\ganttbar[bar/.append style={fill=j4}, bar label font=\color{white} , inline]{$J_5$}{5}{7}
\ganttbar[bar/.append style={fill=j1}, bar label font=\color{white} , inline]{$J_2$}{8}{11}
\ganttbar[bar/.append style={fill=j0}, bar label font=\color{white} , inline]{$J_1$}{12}{14}  
\ganttbar[bar/.append style={fill=j2}, bar label font=\color{black} , inline]{$J_3$}{16}{17}
\ganttbar[bar/.append style={fill=j3}, bar label font=\color{white} , inline]{$J_4$}{18}{19} \\
\ganttbar{$M_4$}{0}{-1}
\ganttbar[bar/.append style={fill=j4}, bar label font=\color{white} , inline]{$J_5$}{8}{8}
\ganttbar[bar/.append style={fill=j1}, bar label font=\color{white} , inline]{$J_2$}{12}{14}
\ganttbar[bar/.append style={fill=j0}, bar label font=\color{white} , inline]{$J_1$}{15}{17}
\ganttbar[bar/.append style={fill=j2}, bar label font=\color{black} , inline]{$J_3$}{18}{18}
\ganttbar[bar/.append style={fill=j3}, bar label font=\color{white} , inline]{$J_4$}{20}{21}  
\end{ganttchart}}
 \caption{Gantt chart of the solution $\pi = \{J_5,J_2,J_1,J_3,J_4\}$ of a 5-jobs 4-machines NWFSP instance. $d_{i,i'}$ is the delay between two consecutive jobs $i$ and $i'$. \label{fig:ganttnwfs}}
 \end{figure}
 
%La figure fournit une représentation d'une solution d'une instance à 4 machines et 5 jobs avec un diagramme de Gantt. Sur cette représentation, le délai entre chaque job est indiqué. On remarque facilement que le Makespan correspond bien à la somme des délais entre les jobs et la somme des processing times du dernier job.

For a better understanding of the specificities of the No-Wait Flowshop Scheduling Problem, Figure~\ref{fig:ganttnwfs} provides a representation of a solution for an instance with five jobs and four machines. %This representation is done with a Gantt chart. 
On this figure, the delay $d_{i,i'}$ between two consecutive jobs $i$ and $i'$ is indicated. 

An interesting property is that the delay is constant and does not depend on the position of the two jobs within the sequence~\citep{Bertolissi_2000}. 
This has a direct impact on the computation of the makespan \Cmax, as it is easy to see on Figure~\ref{fig:ganttnwfs}, which actually corresponds to the sum of the delays between consecutive jobs and all the processing times of the last job scheduled.

Then, the completion time $C_i(\pi)$ of the job $\pi_i$ of sequence $\pi$ can be directly computed from the delays of the preceding jobs as follows:
\begin{equation} 
	C_i(\pi) = \sum^i_{k=2}  d_{\pi_{k-1},\pi_k} + \sum^M_{j=1} p_{\pi_i,j}
	\label{eq:delay_Ci}
\end{equation}	

where $i \in \{2,...,N\}$.
$C_1(\pi)$ is the sum of the processing times on the $M$ machines of the first scheduled job
and then, it is not concerned by the delay.
Therefore, the makespan ($C_{max}=C_{N}(\pi)$) of a sequence $\pi$ can be computed from equation~(\ref{eq:delay_Ci}) with a complexity of $O(N)$.

% Neighborhood

\paragraph{Neighborhood operator} Approaches we will present, involve local search methods.
These ones use a neighborhood operator to move from a solution to another in the search space. 
In this work, we use a permutation representation and the \textit{insertion operator}~\citep{Schiavinotto_2007}, as it is known to make local search more efficient on flowshop problems~\citep{Kouvelis_2000}. 
This operator selects within a sequence $\pi$, a job $\pi_i$ and inserts it at position $k$ ($i \neq k$). 
Hence, jobs between positions $i$ and $k$ are shifted. 
Two sequences $\pi$ and $\pi'$ are said to be neighbors when they differ from exactly one insertion move.
The size of the neighborhood induced by this operator \ie the number of neighboring solutions, is $(n-1)^2$.
It is very interesting to note that exploiting the characteristics of the NWFSP, the makespan of $\pi'$ can be directly computed from the makespan of $\pi$ with a complexity of $O(1)$~\citep{Pan_2007}.

%%%%%%%%%%%%%%%%%%%%%%%%%%%%%%%%%%%%%%%%
%%%%%%%%%%%%%%%%%%%%%%%%%%%%%%%%%%%%%%%%
\subsection{State-of-the-art}
\label{sec:soa}
%%%%%%%%%%%%%%%%%%%%%%%%%%%%%%%%%%%%%%%%
%%%%%%%%%%%%%%%%%%%%%%%%%%%%%%%%%%%%%%%%

%Le NWFSP est NP-Hard. Ainsi, les méthodes exactes ne sont pas capables de trouver une bonne solution en un temps raisonnable pour les problèmes de grandes tailles. C'est pourquoi ces dernières années, plusieurs heuristiques et métaheuristiques ont été développées pour résoudre ce problème.

The NWFSP is NP-Hard when the number of machines is strictly higher than two~\citep{Rock_1984}. 
Thus, exact methods are not able to find the optimal solution in a reasonable time for large-scale instances. 
Recently, several heuristics and metaheuristics have been developed to tackle this problem. 
Table \ref{tab:ref} lists the main approaches of the literature ordered by date and indicates the type of each one, either heuristic (H) or metaheuristic (M).

%Les méthodes heuristiques démarrent généralement d'une séquence initiale ordonnée selon un certain critère. Puis elles construisent une solution en insérant les jobs en optimisant l'objectif à chaque étape. Certaines heuristiques du NWFS sont des adaptations des heuristiques du flowshop de permutation classique et qui s'applique très bien à ce problème. D'autres heuristiques ont été développées... 

\paragraph{Heuristic approaches} Heuristic methods are mostly constructive. They usually start from an initial sequence of jobs ordered according to a criterion. Then, they build a solution by inserting jobs in this order, to optimize the objective function at each step. Some heuristics are either adaptations of heuristics developed for the classical permutation flowshop problem (PFSP) or have been specifically designed for the NWFSP.
Hence, the well-known constructive heuristic NEH \citep{Nawaz_1983}, initially designed for the PFSP, has been successfully applied on the No-Wait variant. Among heuristics specifically designed for the NWFSP, we may cite BIH~\citep{Bianco_1999},  BH~\citep{Bertolissi_2000}, GAN-RAJ~\citep{Gangadharan_1993} and RAJ~\citep{Rajendran_1994}, LC~\citep{Laha_2008}), IBI~\citep{Mousin_2017}.

\paragraph{Metaheuristic approaches} 
Metaheuristics are efficient methods to explore large search space and mostly able to find solutions with a higher quality than constructive heuristics.
Both bio-inspired and \LS~ algorithms have been proposed to tackle the NWFSP. Regarding bio-inspired algorithms, we may find genetic algorithms~\citep{Aldowaisan_2003}, 
particle swarm optimization~\citep{Pan_2008}, or
differential evolution~\citep{Qian_2009}. On the other side, several \LS~ approaches have also been proposed, such as tabu search~\citep{Grabowski_2005,Samarghandi_2012}, variable neighborhood search (VNS)~\citep{Jarboui_2010} or simulated annealing~\citep{Aldowaisan_2003}.
Recently, Ding et al.~\citep{Ding_2015} proposed a very efficient approach named TMIIG (Tabu-Mechanism Improved Iterated Greedy)
based on a variable neighborhood search~\citep{Mladenovic_1997}.
In the perturbation phase, the authors chose to use the efficient destruction-construction method of the Iterated Greedy (IG)~\citep{Ruiz_2007}, initially proposed for the PFSP,  and added a tabu mechanism to avoid scheduling a job at its previous positions during the different destruction-construction phases.

 \begin{table}
	\centerline{
 \begin{tabular}{ccccc}
 	\hline
 	Year & Author(s) & Approach & Type & Ref\\
 	\hline
 	1983 & Nawaz et al. & NEH & H & \citep{Nawaz_1983} \\
 	1993 & Gangadharan et al. & GAN-RAJ & H & \citep{Gangadharan_1993} \\
 	1994 & Rajendran & RAJ & H & \citep{Rajendran_1994} \\
 	1999 & Bianco et al. & BIH & H & \citep{Bianco_1999} \\
 	2000 & Bertolissi & BH & H & \citep{Bertolissi_2000} \\
	2003 & Aldowaisan et al. & GA & M & \citep{Aldowaisan_2003} \\
	2003 & Aldowaisan et al. & SA & M & \citep{Aldowaisan_2003} \\
	2005 & Grabowski et al. & TS & M & \citep{Grabowski_2005} \\	
 	2008 & Laha et al. & LC & H & \citep{Laha_2008} \\
	2008 & Pan et al. & PSO & M & \citep{Pan_2008} \\ 
	2011 & Jarboui et al. & GA-VNS & M & \citep{Jarboui_2010} \\ 
	2012 & Samarghandi et al. & TS-PSO & M & \citep{Samarghandi_2012}\\
	2013 & Davendra et al. & DSOMA & M & \citep{Davendra_2013}\\
	2015 & Ding et al. & TMIIG & M & \citep{Ding_2015} \\
	2017 & Mousin et al. & IBI & H & \citep{Mousin_2017} \\
 \end{tabular}
 }
 	\caption{Heuristics (H) and Meta-heuristics (M) for the NWFSP}
	\label{tab:ref}
 \end{table}

%%%%%%%%%%%%%%%%%%%%%%%%%%%%%%%%%%%%%%%%
%%%%%%%%%%%%%%%%%%%%%%%%%%%%%%%%%%%%%%%%
\subsection{Benchmark}
\label{sec:bench}
%%%%%%%%%%%%%%%%%%%%%%%%%%%%%%%%%%%%%%%%
%%%%%%%%%%%%%%%%%%%%%%%%%%%%%%%%%%%%%%%%
Taillard's benchmark~\citep{Taillard_1993}, initially provided for the PFSP,
is also widely used in the literature for the NWFSP. 
This benchmark proposes 120 instances, organized by 10 instances of 12 different sizes
with a number of jobs $N\in \{20,50,100,200,500\}$ and $M\in \{5,10,20\}$.
The higher the number of jobs and/or the number of machines, 
the more difficult the instance to solve.
Data provided by these instances are the processing times of each job on each machine.
Taillard's instances are said to be random
as the processing times are uniformly generated according to $\mathcal{U}[1;99]$. 
As far as we know, TMIIG is currently one of the best algorithms to solve Taillard's instances
since it has recently (in 2015) found out several new best solutions for the largest instances.

%%%%%%%%%%%%%%%%%%
\section{Super-jobs: Promising Sub-Sequences of Consecutive Jobs}
\label{sec:superjob}
%%%%%%%%%%%%%%%%%%
For the NWFSP, each job is processing without interruption between the successive machines.
Therefore, a question arises: does this specificity lead to a particular structure of the best solutions of a given instance. 
In this section, we conduct an analysis of the global and local optimum solutions in order to extract structural information on them. This analysis leads us to define a promising sub-sequence of consecutive jobs as a {\it super-job}.
Then, we present a methodology to identify {\it super-jobs} of an unknown instance in order to use them to solve it.

%%%%%%%%%%%%
\subsection{Structural Analysis of Optimum Solutions}
%%%%%%%%%%%%

This analysis aims at extracting similarities in the structure of efficient schedules \ie good quality solutions.
We conduct this analysis on \textit{small} instances (low number of jobs) with processing times uniformly generated following the methodology of Taillard's instances (see Section~\ref{sec:bench}). 
We report here, as an example, the analysis of an instance with 12 jobs and 5 machines. This problem size (12) enables to exhaustively enumerate the search space and therefore, to identify the global optimum and the best local optima\footnote{Local optima are solutions that  have no better neighboring solutions \ie no insertion move could lead to a strictly better quality solution.}.
Indeed, local optima are interesting to analyze since they may trap local search methods that explore the search space moving iteratively to improving neighbors.
In the following, the term optimum solutions (or optima) is used to deal with the global or local optimum solutions more generally.

\begin{figure}
\centerline{
\begin{tabular}{r|cccccccccccc}
	$C_{max}$ & \multicolumn{12}{c}{Solution} \\
	\hline
	& \\
	Global Optimum: 1021 	& 	8 	& 	3 	& 	7 	& 	5 	& 	10 	& 	\color{blue}[9 	& 	\color{blue} 1] 	& 	\color{green} [0 	&	\color{green} 4]	& 	6 	& 	\color{red} [11 	& 	\color{red} 2]	\\
	\hline
	& \\
	Local Optima: 1036	&	8	&	3	&	7	&	\color{red} [11	&	\color{red} 2]	&	5	&	10	&	\color{blue} [9	&	\color{blue} 1]	&	\color{green} [0	&	\color{green} 4]	&	6	\\
	1075	&	8	&	10	&	\color{blue}[9	&	\color{blue} 1]	&	\color{green} [0	&	\color{green} 4]	&	7	&	5	&	3	&	\color{red} [11	&	\color{red} 2]	&	6	\\
	1090	&	8	&	3	&	5	&	10	&	6	&	\color{green} [0	&	\color{green} 4]	&	7	&	\color{red} [11	&	\color{red} 2]	&	\color{blue} [9	&	\color{blue} 1]	\\
	1103	&	8	&	3	&	5	&	10	&	6	&	\color{green} [0	&	\color{green} 4]	&	7	&	\color{blue} [9	&	\color{blue} 1]	&	\color{red} [11	&	\color{red} 2]	\\
	1132	&	8	&	10	&	6	&	\color{green} [0	&	\color{green} 4]	&	7	&	5	&	3	&	\color{blue} [9	&	\color{blue} 1]	&	\color{red} [11	&	\color{red} 2]	\\
	1132	&	8	&	3	&	7	&	\color{red} [11	&	\color{red} 2]	&	5	&	10	&	9	&	6	&	\color{green} [0	&	\color{green} 4]	&	1	\\
	$*$1176	&	8	&	3	&	5	&	10	&	\color{blue} [9	&	\color{blue} 1]	&	\color{red} [11	&	\color{red} 2]	&	7	&	6	&	\color{green} [0	&	\color{green} 4]	\\
	1189	&	8	&	10	&	6	&	\color{green} [0	&	\color{green} 4]	&	7	&	5	&	3	&	\color{red} [11	&	\color{red} 2]	&	\color{blue} [9	&	\color{blue} 1]	\\
	1232	&	8	&	10	&	\color{blue} [9	&	\color{blue} 1]	&	\color{green} [0	&	\color{green} 4]	&	7	&	3	&	5	&	6	&	\color{red} [11	&	\color{red} 2]	\\
	1246	&	8	&	3	&	7	&	10	&	\color{blue} [9	&	\color{blue} 1]	&	\color{red} [11	&	\color{red} 2]	&	5	&	6	&	\color{green} [0	&	\color{green} 4]	\\
	%1291	&	8	&	10	&	6	&	\color{green} [0	&	\color{green} 4]	&	7	&	\color{red} [11	&	\color{red} 2]	&	5	&	3	&	\color{blue} [9	&	\color{blue} 1]	\\ 
\end{tabular}
}
	\caption{Description (\Cmax $+$ sequence of jobs) of the global optimum and the 10 best local optima for the studied instance of size 12. The sub-sequences [11~2] and [0~4] colored in red and green respectively, appear in the sequence of all solutions and the sub-sequence [9~1] colored in blue, appears in 10 solutions over 11. When these 3 sub-sequences are considered as 3 unique jobs, one local optimum remains only (identifiable with the star~$*$), the other ones are no longer local and moved to the global optimum.   \label{fig:ig}} 
\end{figure}

%\begin{figure}
%\centerline{
%\begin{tabular}{r|cccccccccccc}
%	$C_{max}$ & \multicolumn{12}{c}{Solution} \\
%	\hline
%	& \\
%	Global Optimum: 1021 	& 	8 	& 	3 	& 	7 	& 	5 	& 	10 	& 	\color{blue}[9 	& 	\color{blue} 1] 	& 	\color{green} [0 	&	\color{green} 4]	& 	6 	& 	\color{red} [11 	& 	\color{red} 2]	\\
%	\hline
%	& \\
%	Local Optima: 1176	&	8	&	3	&	5	&	10	&	\color{blue} [9	&	\color{blue} 1]	&	\color{red} [11	&	\color{red} 2]	&	7	&	6	&	\color{green} [0	&	\color{green} 4]	\\
%\end{tabular}
%}
%	\caption{Quality ($C_{max}$) and description of the global optimum and best local optima obtained considering colored sub-sequences of consecutive jobs are colored as jobs. \label{fig:ig2}} 
%\end{figure}

Figure~\ref{fig:ig} gives the global optimum and the 10 best local optima of the studied instance of size 12. 
For this small instance, it is easy to see that these best optima share a similar structure between them and with the global optimum.
Indeed, job~$8$ is always positioned at the beginning of the schedule and two sub-sequences of two consecutive jobs are present in all of them: $[0~4]$ in green, and $[11~2]$ in red and one sub-sequence in 10 over 11 solutions: $[9~1]$ colored in blue. 
Local optima can not be improved by applying the insertion operator.
However, if we consider each identified sub-sequence of consecutive jobs as a unique job, the best local optimum ($C_{max}=1036$) differs from the global optimum ($C_{max}=1021$) by the single move of the sub-sequence $[11~2]$ only. In the same manner, applying the insertion operator on the other local optima with the consideration of the identified sub-sequences instead of single jobs moves all the local optima (except the one of $C_{max}=1176$ identified with a star) to the global optimum. 

As mentioned before, this study has been conducted on a small instance to be able to exhaustively enumerate the search space. Here, only sub-sequences of two consecutive jobs were found. However, the size is not limited. Hence, if a job $a$ is always followed by a job $b$ and, the job $b$ is always followed by a job $c$ then, the sub-sequence $[a~b~c]$ of three consecutive jobs has to be considered rather than the two sub-sequences $[a~b]$ and $[b~c]$ separately.

Observations made in this analysis motivate the substitution of original jobs by promising sub-sequences of jobs to favor to reach better quality solutions. 
This transformation of the original problem may represent a good opportunity to solve large size instances. 
Therefore a question arises: how to define and identify the promising sub-sequences?

%%%%%%%
\subsection{Definition of Super-Jobs}
%%%%%%%

The previous exhaustive analysis of the structure of local optima for small size instances leads us to suppose that a similar behavior appears on larger ones. In this section, we present the methodology we propose to identify promising sub-sequences of an unknown instance to be solved.

Regardless the size of the search space, good quality solutions and more precisely, good quality local optima, hopefully share a similar structure.
When the search space is non enumerable, it is commonly admitted to use a sample of solutions in order to analyze their structure, characteristics~\dots

Here, we propose to extract the promising sub-sequences from a pool $\mathcal{P^*}$ of \emph{good quality} local optima and to define a \textit{super-job} with a confidence of $\sigma$, any sub-sequence of consecutive jobs that appears at least $\sigma$ percent of times in solutions of $\mathcal{P^*}$.
For example, if $\sigma=50 \%$, the super-jobs are the sub-sequences of consecutive jobs shared by half of the solutions of $\mathcal{P^*}$.
Let us note that only the longest sub-sequences are considered as super-jobs.
For example, if both $[a~b]$ and $[b~c]$ appear at least  $\sigma$ percent of times in $\mathcal{P^*}$, only $[a~b~c]$ is defined as a super-job.

This methodology has the advantage to be relevant regardless the  problem size.
However, the main drawback may be the computational time required to generate the pool of good quality solutions.
Therefore, the size of the pool has to be fixed carefully: too large means that too much time would be spent to generate the pool, too small means that the identification of the super-jobs would be insignificant. This aspect will be discussed during the experimental section.

%%%%%%%
\subsection{Advantages of Super-Jobs}
%%%%%%%

The advantages of considering sub-sequences of consecutive jobs (super-jobs) as unique jobs are several. First, for a complexity point of view, it  will reduce the combinatorics of the problem \ie the number of potential solutions. Secondly, for a local search point of view, this will modify the search space and the  landscape induced by the insertion operator and so, new regions of the search space may become reachable.

%Pour fournir une meilleure compréhension de l'impacte des super-jobs sur le paysage, nous présentons sur la Figure~\ref{fig:landscape-js} la transformation du paysage avec l'utilisation des super-jobs. Chaque point du graphique représentant une solution. En bleu, la représentation sans super-job, en rouge, la représentation avec super-jobs. La représentation avec Super-jobs lisse le paysage ce qui permet d'éviter plus facilement les optima locaux.

\begin{figure}
	\centering
	\includegraphics[scale=0.25]{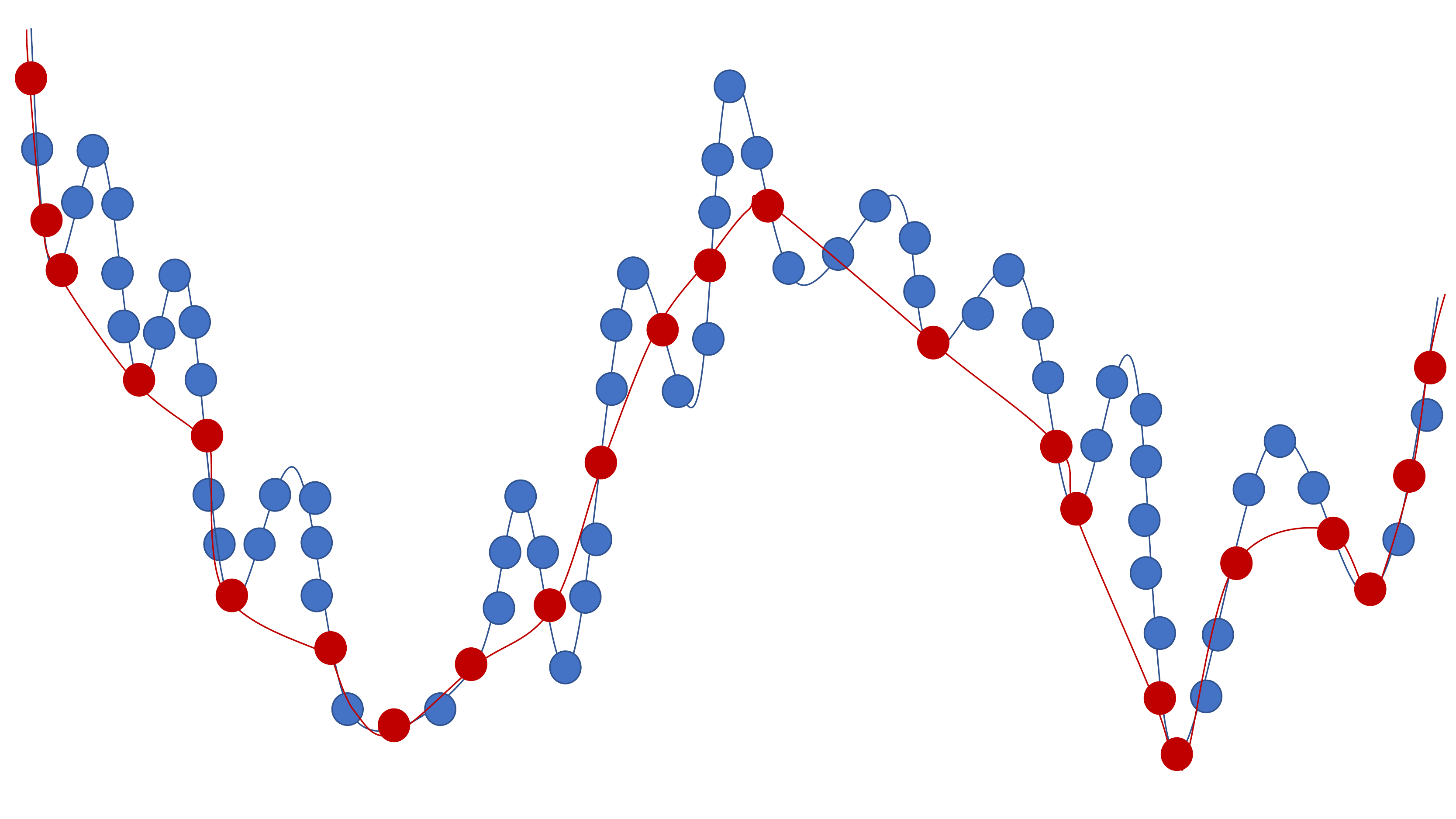} 
	\caption{Visualization of the landscape in 2D. The blue points represent solutions of the original problem while the red ones represent those built with the super-jobs.}
	\label{fig:landscape-js}
\end{figure}

In order to provide a better understanding of the impact of super-jobs on the landscape, we present a visualization in 2D of the landscape transformation in Figure~\ref{fig:landscape-js}. Each point of the graph represents a solution. Blue points represent the original landscape, considering solutions constructed from original jobs following a neighborhood relationship (in the simplified representation, a solution has two neighbors). Red points represent the modified landscape obtained while considering super-jobs. Clearly, the use of super-jobs smooths the landscape and makes it easier to avoid some original local optima which disappeared.

\section{Iterated Greedy with Learning \label{sec:method}}

Super-jobs have been defined as common structural characteristics of good quality solutions. In this section we propose an approach that exploits these super-jobs to improve an already efficient heuristic proposed in the literature for the PFSP -- the Iterated Greedy (IG) -- in order to reach new best solutions for the Taillard's instances. Thus, this section presents the IG Algorithm first, and then the proposed approach.

%%%%%%%%%%
\subsection{Iterated Greedy Algorithm}
%%%%%%%%%%

The Iterated Greedy (IG) algorithm~\citep{Ruiz_2007}, initially proposed for the classical Permutation Flowshop Scheduling Problem (PFSP), is an iterated local search, based on the insertion operator, whose perturbation phase removes some jobs from a solution, and reinserts them one by one at their best position \ie the position that minimizes the partial makespan.
The local search itself  is an iterative improvement: each job of the sequence is considered in a random order and is re-inserted at its best position. 
This process is repeated until a local optimum is reached.
The acceptance criterion of IG is inspired from the one of the simulated annealing and, checks if the new local optimum found is better or not than the best one ever found during the run.
IG is known to be efficient to solve many variants of PFSP.  However, even if it is able to reach good quality solutions in a reasonable computational time, 
it is not able to reach, for the NWFSP, the best-known solutions of the largest instances of Taillard's. 
Indeed, currently, the best algorithm for the NWFSP is the recent algorithm TMIIG~\citep{Ding_2015}, inspired from IG (see Section~\ref{sec:soa}).
Since the performance of IG is doubtless to solve small and medium sizes instances and, since the use of super-jobs decreases the problem size, 
we propose to design a new algorithm taking advantage of both IG and the super-jobs.

%%%%%%%%%%
\subsection{Iterated Greedy with Super-Jobs Algorithm}
%%%%%%%%%%

\begin{algorithm}
	\KwData{$\mathcal{P^*}$: pool of solutions\; 
	\hspace{1cm}$\Sigma = \sigma_1, \sigma_2, \dots $: list of confidence levels in increasing order\;
	\hspace{1cm }$\texttt{SJ}$: list of super-jobs\; 
	\hspace{1cm}$\pi$: solution.}
	
	$\texttt{SJ} =$ identify($\mathcal{P^*}, \sigma_1$) \tcc*{Identifies Super-jobs with a confidence $\sigma_1$ }
	$\pi =$ init($\texttt{SJ}$) \tcc*{Initializes $\pi$ with identified $\texttt{SJ}$}
 	\ForEach{$\sigma$ in $\Sigma$}{
 		$\texttt{SJ} =$ identify($\mathcal{P^*}, \sigma$) \tcc*{Identifies Super-jobs with a confidence $\sigma$ (i)}
 		%\tcc*{Not executed for $\sigma_1$}
 		$\pi =$ IG($\pi,\texttt{SJ}$)\tcc*{Runs IG from $\pi$ with identified $\texttt{SJ}$ (ii)}
 	}
 	\Return{$\pi$}
 	\caption{\IGl -- Iterated Greedy with Learning algorithm.}
 	\label{algo:SJ}
\end{algorithm}

\begin{figure}
	\centering
	\includegraphics[scale=0.4]{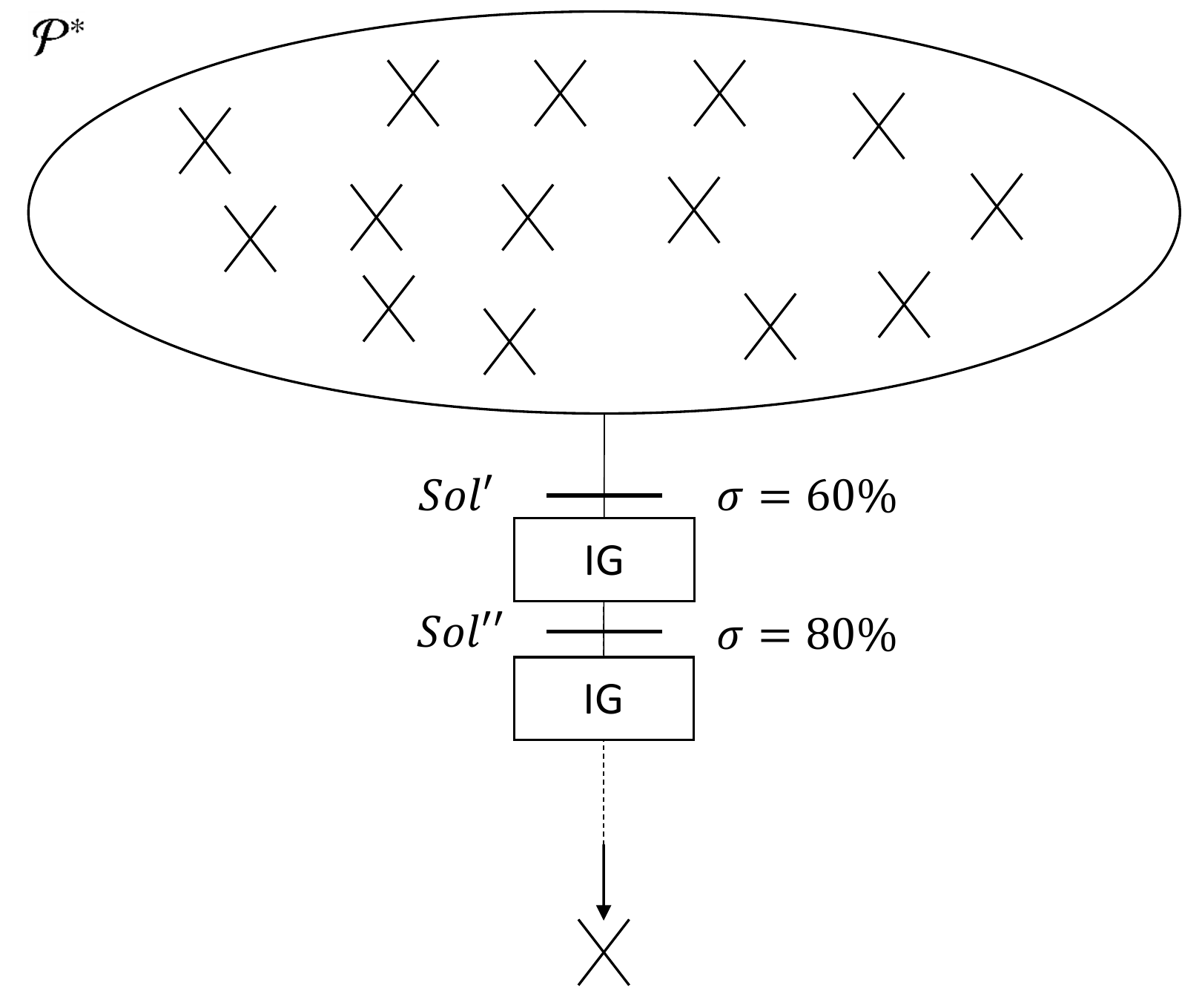} 
	\caption{Illustration of \IGl~ algorithm. }
	\label{fig:SJ-schema}
\end{figure}

The Iterated Greedy with super-jobs algorithm (\IGl) identifies super-jobs of several increasing levels of confidence during the search and exploits them into the basic IG algorithm~\citep{Ruiz_2007}. Algorithm~\ref{algo:SJ} gives the pseudo-code of this new algorithm.
Given a pool of good quality local optima, whose generation will be discussed later, 
and an increasing list $\Sigma = \{\sigma_1, \sigma_2, \dots , \sigma_n\} $ of levels of confidence,
\IGl~ first identifies super-jobs regarding the first level of confidence $\sigma_1$. 
Afterwards, an initialization method generates a first solution $\pi$ with these identified super-jobs.
A process is then, iterated for each level of confidence of $\Sigma$, alternating between (i) a phase of super-jobs identification (except for the first level of confidence $\sigma_1$) and (ii) a phase of improvement using IG. 
The Iterated Greedy algorithm is executed on the solution considering super-jobs as jobs of the problem.  
As IG has no natural stopping criterion, a maximal time, as well as a maximal number of iterations without improvement, are used to stop the IG phase. Once all the levels of confidence have been used, the algorithm returns the best-found solution over the run. Figure~\ref{fig:SJ-schema} illustrates \IGl~  behavior with two levels of confidence $\Sigma=\{60\%,80\%\}$.
%\mek{Figure 4 necessaire ??}

% \cdh{25/08}{Indiquer, entre les 2 valeurs de sigma, que l'on execute IG?}

\begin{table}
	\setlength{\tabcolsep}{2pt}
	\small
	\centerline{
	\begin{tabular}{l|c|c|c|cccccccccccccccccccc}
		$\sigma$ & Phase & Problem size & $C_{max}$ & \multicolumn{20}{c}{Solution $\pi$}\\
		\hline
		60  & (i) & 7  & - &\textbf{[1}&\textbf{7]}&2&\textbf{[3}&\textbf{19]}&5&11&\textbf{[13}&\textbf{15}&\textbf{14}&\textbf{17}&\textbf{9}&\textbf{4}&\textbf{8}&\textbf{18}&\textbf{0}&\textbf{12}&\textbf{6}&\textbf{10]}&16\\
		 & \emph{init} &  &3021&\textbf{[3}&\textbf{19]}&16&5&2&11&\textbf{[13}&\textbf{15}&\textbf{14}&\textbf{17}&\textbf{9}&\textbf{4}&\textbf{8}&\textbf{18}&\textbf{0}&\textbf{12}&\textbf{6}&\textbf{10]}&\textbf{[1}&\textbf{7]}\\
		& (ii) &  &3021&\textbf{[3}&\textbf{19]}&16&5&2&11&\textbf{[13}&\textbf{15}&\textbf{14}&\textbf{17}&\textbf{9}&\textbf{4}&\textbf{8}&\textbf{18}&\textbf{0}&\textbf{12}&\textbf{6}&\textbf{10]}&\textbf{[1}&\textbf{7]}\\
		\hline
		80   & (i) & 14 &3021&\textbf{[3}&\textbf{19]}&16&5&2&11&\textbf{[13}&\textbf{15]}&14&17&\textbf{[9}&\textbf{4]}&8&\textbf{[18}&\textbf{0]}&12&\textbf{[6}&\textbf{10]}&\textbf{[1}&\textbf{7]}\\
		 & (ii) &  &3013&\textbf{[3}&\textbf{19]}&16&5&2&12&17&\textbf{[18}&\textbf{0]}&11&\textbf{[13}&\textbf{15]}&14&\textbf{[9}&\textbf{4]}&8&\textbf{[6}&\textbf{10]}&\textbf{[1}&\textbf{7]}\\
		\hline
		$\infty$ & (i) & 20 &3013&3&19&16&5&2&12&17&18&0&11&13&15&14&9&4&8&6&10&1&7\\
		 & (ii) &  & 3013&3&19&16&5&2&12&17&18&0&11&13&15&14&9&4&8&6&10&1&7\\
	\end{tabular}
	}
	\caption{\IGl~  in the different phases on instance \texttt{ta023} of Taillard with $\Sigma=\{60\%;70\%;\infty\}$. Phase (i) corresponds to the identification of super-jobs and phase (ii), to the application of IG. For each confidence level, the identified super-jobs are in bold.}
    	\label{tab:SJevo}
\end{table}

Table~\ref{tab:SJevo} presents the evolution of the solution $\pi$ for all phases of an execution of \IGl~  on the instance \texttt{ta023} of Taillard (20 jobs, 20 machines)
with the list of confidence levels $\Sigma=\{60\%,80\%,\infty\}$, where $\sigma=\infty$ means that no super-job is created (the problem is solved with all the original jobs). 
In this example, with $\sigma=60\%$, seven super-jobs are identified (phase (i)):
one of size 12, two of size 2 and the fourth-remaining ones of size 1. 
Therefore, the problem size is decreased from 20 to 7.
The initialization method then builds a local optimum with a quality of 3021
that IG is not able to improve (phase (ii)). 
Then, considering a confidence level of $\sigma=80\%$, some previous super-jobs are decomposed (phase (i)).
Indeed, the largest super-job is decomposed into eight smaller ones.
The problem size is equal to 14 and IG manages to find a better solution with a quality of 3013 (phase (ii)). 
It appears, in this special case, that the global optimum is reached within this second phase. 
This explains why the last phase ($\sigma=\infty$, super-jobs are all of size 1) is not able to produce any improving solution.
%%%%%%%
\section{Experiments \label{sec:experiments}}

In order to assess the efficiency of the \IGl~  algorithm, experiments are driven on Taillard's benchmarks and results compared to the previous best-known solutions of the literature obtained by TMIIG~\citep{Ding_2015}.

\subsection{Experimental Protocol \label{sec:protocol}}

As exposed in section~\ref{sec:soa}, the benchmark used to evaluate the performance of the proposed method is composed of Taillard's instances~\citep{Taillard_1993} organized by 10 instances of 12 different sizes.
%Instances with $20$ jobs are not considered here as optimal solutions have already been found in the literature. 
%Moreover, we are more interested in large instances that are more difficult to solve.
The algorithm was implemented in C++ and the experiments were executed on an Intel(R) Xeon(R) 3.5GHz processor.\\

Following a preliminary study, several parameters was settled for these experiments: 
\begin{itemize}
\item {\bf Initial pool of solutions $\mathcal{P}^*$:} 10 solutions (enough to extract knowledge) were generated from 10 independent executions of IG with a maximal time of $ n ^ 2 * 10$  ms each.
\item {\bf Levels of confidence:} Two lists $\Sigma_1= \{60\%,80\%,\infty\}$ and $\Sigma_2=\{60\%,70\%, 80\%, 90\%, \infty\}$ were tested in order to evaluate the performance of \IGl~ according to the granularity.
\item {\bf Initialization method:} Iterated Best Insertion heuristic~\citep{Mousin_2017} a constructive heuristic hybridized with a basic local search (first improvement hill climbing).
%The designed initialization procedure (used at phase 1 for  $\sigma_1$) is inspired from NEH~\citep{Nawaz_1983}, an efficient constructive heuristic, hybridized with a stochastic local search to reach local optima.
\item {\bf Stopping criterion of IG in phase ii:} A maximal time of $n_{sj} ^2 * 10$ ms (where $n_{sj}$ is the number of super-jobs of the phase), and a maximal number of iterations without improvement of $50 * n_{sj}$ are defined.
\end{itemize}

Each execution of \IGl~  on a given instance $\mathcal{I}$ returns a solution $\pi$ of fitness $C_{max}(\pi)$. 
To measure the quality of the solution, the Relative Percentage Deviation (RPD) is computed relatively to the best-known solution of the literature $\pi^*$ 
%% ??of $\mathcal{I}$ 
as follows:
\begin{equation}
	RPD = \frac{C_{max}(\pi) - C_{max}(\pi^*)}{C_{max}(\pi^*)} * 100
\end{equation}
Hence a negative RDP indicates that a new best known solution is found.

\IGl~  is stochastic, thus 30 runs were executed to make the experimental results robust, and performance for an instance $\mathcal{I}$ is the average of the 30 RPD computed.

%%%%%%
\subsection{Experimental Results\label{sec:experimentalRes}}

\begin{table}
\setlength{\tabcolsep}{2pt}
\resizebox{\textwidth}{!}{
\begin{tabular}{|c|*{40}{c|}}
\hline
\diagbox{Machines}{Jobs} & \multicolumn{10}{c|}{\texttt{50}} & \multicolumn{10}{c|}{\texttt{100}} & \multicolumn{10}{c|}{\texttt{200}} & \multicolumn{10}{c|}{\texttt{500}} \\\hline

\RowColor \texttt{5} & 

~~ & ~~ & \scriptsize 17 & \scriptsize 2~ & ~~ & \scriptsize 21 & ~~ & ~~ & \scriptsize 11 & ~~ & 

\scriptsize X & \scriptsize 14 & \scriptsize 5 & \scriptsize 29 & \scriptsize 13 & \scriptsize 11 & \scriptsize 16 & \scriptsize 5 & \scriptsize 6 & \scriptsize 4 &

\multicolumn{10}{c|}{\cellcolor{white} ~} & \multicolumn{10}{c|}{\cellcolor{white} ~} \\

\RowColor \texttt{10} & 

~ & ~ & ~ & ~ & ~ & ~ & ~ & ~ & ~ & ~ & 

\scriptsize 26 & \scriptsize 20 & \scriptsize 7 & \scriptsize 24 & \scriptsize 18 & \scriptsize 26 & \scriptsize 22 & \scriptsize X & \scriptsize 23 & \scriptsize 23 &

\scriptsize 24 & \scriptsize 19 & \scriptsize X  & \scriptsize X & \scriptsize X & \scriptsize X & \scriptsize 29 & \scriptsize 28 & \scriptsize X & \scriptsize X &

 \multicolumn{10}{c|}{\cellcolor{white} ~} \\

\RowColor \texttt{20} & 

~ & ~ & ~ & ~ & ~ & ~ & ~ & ~ & ~ & ~ & 

\scriptsize 13 & \scriptsize 29 & \scriptsize 12 & \scriptsize 25 & \scriptsize 26 & \scriptsize X & \scriptsize 29 & \scriptsize X & \scriptsize 19 & \scriptsize 29 &

\scriptsize X & \scriptsize X & \scriptsize X & \scriptsize X & \scriptsize X & \scriptsize X & \scriptsize X & \scriptsize X & \scriptsize 29 & \scriptsize X &

\scriptsize X & \scriptsize X & \scriptsize X & \scriptsize 28 & \scriptsize 29 & \scriptsize 23 & \scriptsize X & \scriptsize X & \scriptsize X & \scriptsize X\\
\hline
\end{tabular}}
\caption{Result on Taillard's instances (organized by size) for $\Sigma_1= \{60\%,80\%,\infty\}$. Gray cell: the best-known solution of the literature is reached. \#/X : Number of times the best-known is improved (X = all runs).}
\label{tab:res1}
\end{table}

 %\cdh{06/09}{Enlever les "cases" dans les tableaux la ou il n'existe pas de benchmarks (200x5, 500x5, 500x10)}

\begin{table}
\setlength{\tabcolsep}{2pt}
\resizebox{\textwidth}{!}{
\begin{tabular}{|c|*{40}{c|}}
\hline
\diagbox{Machines}{Jobs} & \multicolumn{10}{c|}{\texttt{50}} & \multicolumn{10}{c|}{\texttt{100}} & \multicolumn{10}{c|}{\texttt{200}} & \multicolumn{10}{c|}{\texttt{500}} \\\hline

\RowColor \texttt{5} & 

\scriptsize 1~ & ~~ & \scriptsize 23 & \scriptsize 2~ & ~~ & \scriptsize 25 & ~~ & ~~ & \scriptsize 17 & ~~ & 

\scriptsize X & \scriptsize 14 & \scriptsize 6 & \scriptsize X & \scriptsize 13 & \scriptsize 14 & \scriptsize 20 & \scriptsize 8 & \scriptsize 6 & \scriptsize 6 &

\multicolumn{10}{c|}{\cellcolor{white} ~} & \multicolumn{10}{c|}{\cellcolor{white} ~} \\

\RowColor \texttt{10} & 

~ & ~ & ~ & ~ & ~ & ~ & ~ & ~ & ~ & ~ & 

\scriptsize 27 & \scriptsize 17 & \scriptsize 19 & \scriptsize X & \scriptsize 24 & \scriptsize 28 & \scriptsize 26 & \scriptsize X & \scriptsize X & \scriptsize 22 &

\scriptsize 27 & \scriptsize 19 & \scriptsize X  & \scriptsize X & \scriptsize X & \scriptsize X & \scriptsize X & \scriptsize 29 & \scriptsize X & \scriptsize X &

\multicolumn{10}{c|}{\cellcolor{white} ~}  \\

\RowColor \texttt{20} & 

~ & ~ & ~ & ~ & ~ & ~ & ~ & ~ & ~ & ~ & 

\scriptsize 17 & \scriptsize X & \scriptsize 22 & \scriptsize 26 & \scriptsize 27 & \scriptsize X & \scriptsize X & \scriptsize X & \scriptsize 27 & \scriptsize 29 &

\scriptsize X & \scriptsize X & \scriptsize X & \scriptsize X & \scriptsize X & \scriptsize X & \scriptsize X & \scriptsize X & \scriptsize X & \scriptsize X &

\scriptsize X & \scriptsize X & \scriptsize X & \scriptsize X & \scriptsize 29 & \scriptsize 26 & \scriptsize X & \scriptsize X & \scriptsize X & \scriptsize X\\
\hline
\end{tabular}}
\caption{Result on Taillard's instances (organized by size) for $\Sigma_2=\{60\%,70\%, 80\%, 90\%, \infty\}$. Gray cell: the best-known solution of the literature is reached. \#/X : Number of times the best-known is improved (X = all runs).}
\label{tab:res2}
\end{table}

%\begin{figure}[h!]
%	\centerline{
%		\includegraphics[width=1\textwidth,clip=true]{results.png}
%	}
%    \caption{Results on Taillard's instances (organized by size).}
%    \label{fig:Results}
%\end{figure}

To analyze performance of the \IGl~  algorithm, results obtained are compared with the best-known solutions of the literature, reported in~\citep{Ding_2015}.
Tables~\ref{tab:res1} (list $\Sigma_1$) and \ref{tab:res2} (list $\Sigma_2$) indicate, for each instance (10 instances per size) if the best-known solution of the literature is reached (cells colored in gray) and if this best-known is improved (non empty cell). Hence, when a number is present in a cell, this indicates the number of times the method improves the previous best-known solution over the 30 executions, an 'x' indicates 30/30. 
The results obtained for the 30 instances with 20 jobs are not reported here as a simple IG manages to find the optimal solution.
%the optimal solutions have been already found by a simple IG.
Both tables show that for all instances the method reaches the best-known of the literature regardless the two lists of confidence levels $\Sigma_1$ and $\Sigma_2$. In addition they show that all instances with 100, 200 and 500 jobs are improved with the proposed \IGl~  algorithm. 
The detail of the new best makespan values reached by the method is given in Tables~\ref{tab:bestKnownConfig1} and~\ref{tab:bestKnownConfig2} in the~\ref{sec:annexe}.
%\footnote{Best up to date solutions of Taillard's instances are reported in http://lucienmousin.fr/nwfs}.

%%%%
%\paragraph{Analysis}

To deeper analyze the behavior of the method, Tables~\ref{tab:combiAndConv-simple} and \ref{tab:combiAndConv-simple4} report some information about the execution of the method for the two lists  $\Sigma_1$ and $\Sigma_2$. Results presented are average over the 30 executions of all instances (10) of a same size.
Both tables have the same structure and the conclusions are also similar. We will first discuss about the common analyses and then point out differences within the discussion.

%Table~\ref{tab:combiAndConv-simple} reports some information to deeper analyze the behavior of the method in the different phases ($Init$, $IG$) according to the level of confidence. results presented are average over all instances of a same size.

%Both tables have the same structure, the only difference is the number of levels of confidence that are considered: $\sigma=60\%$ and $80\%$ for Table~\ref{tab:combiAndConv-simple};  $\sigma=60\% , 70\%, 80\%$ and $90\%$ for Table~\ref{tab:combiAndConv-simple4}.  As we will see conclusions are also similar. We will first discuss about the common analyses and then point out differences within the discussion.

Left parts of these tables report the size of the problem \ie the number of jobs for each phase. 
These jobs are either original jobs, or the super-jobs constructed by the concatenation of several jobs, as explained before. 
This measure gives the combinatorics of the problem. 
For example, in Table~\ref{tab:combiAndConv-simple} for instances of size 200, when $\sigma=60\%$, the number of jobs is around 80-90. 
This means that the size of the problem has been divided by more than 2. 
In the following phase, when $\sigma=80\%$, some super-jobs are decomposed and the number of jobs is around 130-140. 
The combinatorics is still reduced. 
As mentioned before when $\sigma=\infty$ the number of jobs equals the original number of jobs (\ie 200 for the previous example).
This first observation indicates that there is a real difference between the set of jobs obtained for the two levels of confidence, which shows  that the identification of super-jobs is different.
%as the number of jobs to consider is different. %(which shows that the identification of super-jobs is different).

Another observation is that the number of machines has also an impact on the identification of super-jobs.
Indeed for a given number of jobs, the more the number of machines, the smaller the combinatorics.
As far as instances of size 500 are concerned, the high level of combinatorics may be explained by the use of IG to generate the pool of solutions. 
Indeed, IG has difficulty to converge for large size problems within the time allowed. 
Hence the solutions of the pool are too diversified to identify common sub-sequences of jobs.%good job associations.

\begin{table}[h!]
\resizebox{\textwidth}{!}{
\small
    \begin{tabular}{l|cc||cccc||cccc||cccc}
Instances &
\multicolumn{2}{c||}{Problem size} & 
\multicolumn{4}{c||}{End of improvement} &
\multicolumn{4}{c||}{RPD value}  &
\multicolumn{4}{c}{Time (s)}  \\
                &       60\%&   80\%&   Init&  IG$_{60\%}$&    IG$_{80\%}$&    IG$_{\infty}$&  Init&
IG$_{60\%}$&    IG$_{80\%}$&    IG$_{\infty}$&  60\%&   80\%&    $\infty$ &        total \\
                \hline
\texttt{20$\times$5}&   1.23&   10.26&  98.7&   0.0&    1.3&    0.0&    0.06&  0.06&  0.00&  0.00&  0.14&  0.31& 0.00&  0.45   \\
\texttt{20$\times$10}&  1.00&   10.00&  100.0&  0.0&    0.0&    0.0&    0.00&  0.00&  0.00&  0.00&  0.14&  0.30& 0.00&  0.45   \\
\texttt{20$\times$20}&  1.26&   10.35&  95.3&   2.3&    2.3&    0.0&    0.03&  0.03&  0.00&  0.00&  0.14&  0.30& 0.01&  0.45   \\
\texttt{50$\times$5}&   9.23&   31.65&  10.0&   31.0&   53.7&   5.3&    0.97&  0.22&  0.06&  0.06&  0.59&  1.72& 1.69&  3.99   \\
\texttt{50$\times$10}&  5.98&   29.83&  28.3&   6.7&    57.7&   7.3&    0.72&  0.47&  0.07&  0.07&  0.82&  1.84& 0.98&  3.64   \\
\texttt{50$\times$20}&  4.10&   27.85&  59.0&   3.3&    29.3&   8.3&    0.42&  0.24&  0.05&  0.04&  0.89&  1.88& 0.54&  3.30   \\
\texttt{100$\times$5}&  45.46&  78.32&  0.0&    43.3&   34.7&   22.0&   3.00&  0.03&  0.00&  -0.01& 6.88&  7.07& 8.98&  22.93  \\
\texttt{100$\times$10}& 29.04&  68.99&  0.0&    16.3&   53.7&   30.0&   1.69&  0.07&  -0.05& -0.07& 3.25&  7.48& 9.38&  20.11  \\
\texttt{100$\times$20}& 27.37&  68.44&  0.0&    9.3&    59.0&   31.7&   1.49&  0.09&  -0.08& -0.10& 2.82&  7.71& 9.22&  19.74  \\
\texttt{200$\times$10}& 95.36&  157.41& 0.0&    15.3&   45.7&   39.0&   2.89&  -0.13& -0.18& -0.19& 30.94& 35.48&46.51& 112.93 \\
\texttt{200$\times$20}& 81.83&  150.74& 0.0&    5.0&    41.7&   53.3&   1.99&  -0.20& -0.30& -0.32& 22.12& 39.35&48.80& 110.27 \\
\texttt{500$\times$20}& 268.87& 409.28& 0.0&    0.7&    30.3&   69.0&   2.81&  -0.18& -0.24& -0.25& 598.84& 656.63&        784.92&        2040.38        \\
                \end{tabular}
                }
        \caption{\IGl~  with $\Sigma_1=\{60\%,80\%,\infty\}$. Reported measures are average over the 10 instances of each size.} \label{tab:combiAndConv-simple}
\end{table}

\begin{table}[h!]
\resizebox{\textwidth}{!}{
\small
  \begin{tabular}{l|cccc||cccccc||cccccc||cccccc}
Instances &\multicolumn{4}{c||}{Problem size} & \multicolumn{6}{c||}{End of improvement} &
\multicolumn{6}{c||}{RPD value} &  \multicolumn{6}{c}{Time (s)}  \\
                &       60\%&   70\%&   80\%&   90\%&   Init&  IG$_{60\%}$&    IG$_{70\%}$&    IG$_{80\%}$&
IG$_{90\%}$&    IG$_{\infty}$&  Init&  IG$_{60\%}$&    IG$_{70\%}$&    IG$_{80\%}$&    IG$_{90\%}$&    IG$_{\in
fty}$&  60\%&   70\%&   80\%&   90\%&    $\infty$ &        total \\
                \hline
 \texttt{20$\times$5} 	 & 	 1.23   	 & 	 10.19   	 & 	 10.30   	 & 	 10.20   	 & 	 98.70   	 & 	 0.00     	 & 	 1.33   	 & 	 0.00     	 & 	 0.00     	 & 	 0.00     	 & 	 0.06   	 & 	 0.06   	 & 	 0.00     	 & 	 0.00     	 & 	 0.00     	 & 	 0.00     	 & 	 0.14   	 & 	 0.14   	 & 	 0.14   	 & 	 0.31   	 & 	 0.00     	 & 	 0.74   	 \\ 
 \texttt{20$\times$10} 	 & 	 1.00   	 & 	 10.00   	 & 	 10.00   	 & 	 10.00   	 & 	 100.00   	 & 	 0.00     	 & 	 0.00     	 & 	 0.00     	 & 	 0.00     	 & 	 0.00     	 & 	 0.00     	 & 	 0.00     	 & 	 0.00     	 & 	 0.00     	 & 	 0.00     	 & 	 0.00     	 & 	 0.14   	 & 	 0.14   	 & 	 0.14   	 & 	 0.31   	 & 	 0.00     	 & 	 0.73   	 \\ 
 \texttt{20$\times$20} 	 & 	 1.26   	 & 	 10.22   	 & 	 10.40   	 & 	 10.20   	 & 	 95.30   	 & 	 2.30   	 & 	 1.67   	 & 	 0.67   	 & 	 0.00     	 & 	 0.00     	 & 	 0.03   	 & 	 0.03   	 & 	 0.00   	 & 	 0.00     	 & 	 0.00     	 & 	 0.00     	 & 	 0.14   	 & 	 0.14   	 & 	 0.14   	 & 	 0.30   	 & 	 0.01   	 & 	 0.74   	 \\ 
 \texttt{50$\times$5} 	 & 	 9.23   	 & 	 29.51   	 & 	 31.70   	 & 	 29.50   	 & 	 9.30   	 & 	 23.70   	 & 	 35.67   	 & 	 20.67   	 & 	 9.00   	 & 	 1.67   	 & 	 0.97   	 & 	 0.22   	 & 	 0.06   	 & 	 0.04   	 & 	 0.03   	 & 	 0.03   	 & 	 0.54   	 & 	 1.40   	 & 	 1.40   	 & 	 1.56   	 & 	 1.70   	 & 	 6.60   	 \\ 
 \texttt{50$\times$10} 	 & 	 5.98   	 & 	 28.09   	 & 	 29.90   	 & 	 28.10   	 & 	 24.30   	 & 	 5.30   	 & 	 30.00   	 & 	 19.67   	 & 	 14.67   	 & 	 6.00   	 & 	 0.72   	 & 	 0.47   	 & 	 0.09   	 & 	 0.07   	 & 	 0.06   	 & 	 0.06   	 & 	 0.76   	 & 	 1.29   	 & 	 1.31   	 & 	 1.77   	 & 	 0.99   	 & 	 6.11   	 \\ 
 \texttt{50$\times$20} 	 & 	 4.10   	 & 	 26.91   	 & 	 27.90   	 & 	 26.90   	 & 	 56.70   	 & 	 3.00   	 & 	 14.00   	 & 	 13.00   	 & 	 10.67   	 & 	 2.67   	 & 	 0.42   	 & 	 0.24   	 & 	 0.06   	 & 	 0.04   	 & 	 0.04   	 & 	 0.03   	 & 	 0.86   	 & 	 1.14   	 & 	 1.19   	 & 	 1.87   	 & 	 0.54   	 & 	 5.60   	 \\ 
 \texttt{100$\times$5} 	 & 	 45.46   	 & 	 71.38   	 & 	 78.30   	 & 	 71.40   	 & 	 0.00     	 & 	 31.00   	 & 	 21.33   	 & 	 16.00   	 & 	 19.67   	 & 	 12.00   	 & 	 3.00   	 & 	 0.03   	 & 	 0.00   	 & 	-0.01   	 & 	-0.02   	 & 	-0.03   	 & 	 6.82   	 & 	 6.38   	 & 	 6.94   	 & 	 7.63   	 & 	 8.91   	 & 	 36.68   	 \\ 
 \texttt{100$\times$10} 	 & 	 29.04   	 & 	 63.72   	 & 	 69.00   	 & 	 63.70   	 & 	 0.00     	 & 	 9.00   	 & 	 25.33   	 & 	 26.67   	 & 	 25.33   	 & 	 13.67   	 & 	 1.69   	 & 	 0.07   	 & 	-0.04   	 & 	-0.07   	 & 	-0.09   	 & 	-0.09   	 & 	 3.32   	 & 	 6.57   	 & 	 6.80   	 & 	 7.01   	 & 	 8.92   	 & 	 32.62   	 \\ 
 \texttt{100$\times$20} 	 & 	 27.37   	 & 	 62.91   	 & 	 68.50   	 & 	 62.90   	 & 	 0.00     	 & 	 8.00   	 & 	 16.33   	 & 	 32.00   	 & 	 28.67   	 & 	 15.00   	 & 	 1.49   	 & 	 0.09   	 & 	-0.05   	 & 	-0.10   	 & 	-0.12   	 & 	-0.13   	 & 	 2.90   	 & 	 6.95   	 & 	 7.11   	 & 	 7.01   	 & 	 8.63   	 & 	 32.60   	 \\ 
 \texttt{200$\times$10} 	 & 	 95.36   	 & 	 144.20   	 & 	 157.50   	 & 	 144.20   	 & 	 0.00     	 & 	 5.70   	 & 	 19.33   	 & 	 28.00   	 & 	 23.00   	 & 	 24.00   	 & 	 2.89   	 & 	-0.13   	 & 	-0.18   	 & 	-0.20   	 & 	-0.22   	 & 	-0.22   	 & 	 31.09   	 & 	 30.83   	 & 	 32.45   	 & 	 35.71   	 & 	 43.83   	 & 	 173.91   	 \\ 
 \texttt{200$\times$20} 	 & 	 81.83   	 & 	 138.14   	 & 	 150.90   	 & 	 138.10   	 & 	 0.00     	 & 	 0.70   	 & 	 14.67   	 & 	 28.00   	 & 	 28.67   	 & 	 28.00   	 & 	 1.99   	 & 	-0.20   	 & 	-0.28   	 & 	-0.32   	 & 	-0.33   	 & 	-0.34   	 & 	 22.09   	 & 	 34.00   	 & 	 33.18   	 & 	 35.29   	 & 	 45.30   	 & 	 169.85   	 \\ 
 \texttt{500$\times$20} 	 & 	 268.87   	 & 	 376.81   	 & 	 409.50   	 & 	 376.80   	 & 	 0.00     	 & 	 0.00     	 & 	 5.33   	 & 	 15.67   	 & 	 31.33   	 & 	 47.67   	 & 	 2.81   	 & 	-0.18   	 & 	-0.23   	 & 	-0.26   	 & 	-0.27   	 & 	-0.28   	 & 	 601.14   	 & 	 566.95   	 & 	 585.10   	 & 	 631.06   	 & 	 744.54   	 & 	 3 128.79   	

        \\
                \end{tabular}
                }
        \caption{\IGl~  with $\Sigma_2=\{60\%,70\%, 80\%, 90\%, \infty\}$. Reported measures are averages over the 10 instances of each size.}

        \label{tab:combiAndConv-simple4}
\end{table}

Measures about the convergence of \IGl~  are given in the middle part of the two tables.
The column, called \emph{End of improvement}, indicates in percentage, the number of times (over the 30 executions) the method -- $Init$ or $IG$ -- reaches the best solution of the run (average of the 10 instances of each size). 
We can observe, that in both tables, for instances of size 20, the best solution is mainly reached during the initialization phase. 
Indeed, for small size instances, IG is very efficient and manages to reach best-known solutions. 
For size 50, the \IGl~  manages to almost always find its best solution before considering the original problem  ($\sigma=\infty$)
contrary to the original IG. This validates the use of super-jobs.
However, for largest size problems, some improvements are still obtained in the last phase when the original problem is considered. %, and the initialization is quite far from best solutions.
%Up to size 100 the three first phases are sufficient to reach very good solutions.
%The super-jobs identified with $\sigma=70\%$ are pertinent and, the combinatorics has been highly decreased.
%For larger size problems, some improvements are still obtained in the last phase when the original problem is considered.
%We could wonder whether an intermediate phase with $\sigma=80\%$ could help to find out better or best solutions.

This analysis is re-enforced by the third set of columns of both tables that report the average RPD at each phase. 
Thus it indicates, how far from the best-known solution of the literature, are solutions reached after each phase. Let us recall that a negative value indicates that a new best solution has been found.
For instances of size 20, the best-known solution, which is optimal, is reached (RPD=0). For size 50, the best-known solutions  are often reached (but maybe not in all the executions, which explained a small positive RPD). 
The most interesting is for large instances, as best known solutions are improved (negative RPD). A complementary interesting observation for large size instances is the high improvement between the initialization and the first phase with $\sigma=60\%$. In the following phases, even if some better solutions are reached, the improvement is less important but significant.

The right part of the two tables reports time spent at each phase. The stopping criterion, at each phase, is either a maximal time or a maximal number of iterations without improvements, both depending on the number of (super-)jobs (see Section \ref{sec:protocol}). So, as we can expect, for most cases, the time spent increases with the value of confidence level. Note that the total time is quite significant (the computational time for the generation of the pool of solutions still have to be added), but the objective of the approach is to be able to find new best solutions. So we do not consider real computing time constraints.

%%%%%%%%%%%%
\subsection{Discussion}
%%%%%%%%%%%%

The experimental results proved the performance of our approach (\IGl) since new best solutions have been found out for every largest Taillard's instance.
This section provides a discussion on two important aspects of the method: the computational time required for the generation of the pool of solutions, and the analysis of the dynamic of the method.

%%%%
\paragraph{Discussion on the generation of the pool $\mathcal{P^*}$}

The computational time of the proposed method may represent a drawback when solving large instances. This computational time is partly explicated by the generation of the pool of solutions $\mathcal{P^*}$.

The fact is that the performance of the approach is based on the knowledge extraction from this pool of initial solutions used to identify pertinent super-jobs.
%The computational time of the generation of this pool increases with the number of jobs and machines of the instance.
Preliminary results show that the quality of the solutions of the pool impacts a lot the performance, and hence constructive heuristics do not give enough good quality solutions to identify reliable super-jobs. 
Hence, we chose IG, with a time limit, to give pretty good quality solutions for the pool of solutions, but it is time consuming.
In addition, we tested different sizes for the pool. Indeed, the higher the size, the larger the computational time to generate the pool whereas the lower the size, the lower the chance to have a representative pool.
Following these tests we decided to generate a pool of 10 solutions only, as pertinent super-jobs can be found out, even with so few solutions. A small pool reduces a lot the whole computational time of the approach. However, the time still remains important. For example, to solve a 200 jobs instance, 400 seconds are required to generate one solution of the pool, hence around one hour for the 10 solutions of the pool.  This is quite long, but not so important if we want to obtain new best solutions.

Since our approach is stochastic, in the exposed experiments, the performance is evaluated from 30 executions of \IGl~ for each instance. Each one generates its own pool of good quality solutions. 
We drove some parallel experiments where a single pool of 10 solutions was generated only and was shared between the 30 executions. 
The experimental results were similar: best-known solutions of the literature were reached for the smallest instances of Taillard, and improved for the largest one. 
Using such a shared pool decreases the whole computational time for the 30 executions.
%The computational time of the method may be decrease with this shared pool.

During the analysis of the method, we also noticed that for largest instances (with 500 jobs, mainly) a better pool of solutions improves the identification of super-jobs and then still reduces the combinatorics.
Thus, we can imagine leaving more computational time to IG to generate a shared pool of better quality solutions, instead of generating one different pool for each execution.

%the structure of the solutions of the pool for the no-wait flowshop scheduling problem.
%These experiments enabled to strengthen the interest of exploiting the extracted knowledge from 

%%%%
\paragraph{Discussion on the behavior of the method}
Another interesting aspect is that the performance of the approach lies on the reduction of the combinatorics of the problem made possible by the particular structure of the best quality local optima.
In the search space, each local optimum is the 'center' of a basin of attraction. 
All the basins make the landscape very rugged for \LS~  methods.
The perturbation phase of IG has been designed to escape from local optima, and so, from their attraction basins.
However, the basins of attraction are not side-by-side but included in each other;  the best local optimum of a large basin may be the center of other ones. 
Hence, even with a perturbation, a \LS~  method often remains in the same large basin of attraction it started. 
The reduction of the combinatorics of the problem, with the identification of super-jobs, produces an interesting effect on the landscape.
Indeed, some original local optima do not exist in the reduced landscape and so, for its  basins of attraction.
Therefore, regions of the landscape are smoothed,  original basins of attractions get larger and, the performance of IG at each iteration of our approach is improved.  
For example, for instances with 20 jobs (the easiest of the Taillard instances), IG ends to converge close to the best-known solutions without reaching it, whereas with the reduction of the combinatorics, it reaches it each time.
The reduction of the number of  basins of attraction helps IG to move towards the best one. 
Exploiting super-jobs erases rugged regions of the search space and increases the performance of IG. 
These encouraging results lead us to incorporate  (\IGl) in a more general scheme.

\section{\IIGl : an Iterative Version \label{sec:iteratedsuperjob}}

\subsection{Description}

The experimental results presented above, showed that the learning mechanism is efficient for both small and large size instances.
Indeed, \IGl~ is able to either find out new best solutions 
or at least to reach the best-known solutions for the Taillard instances. 
For the largest (and actually the most difficult) instances of size 100, 200 and 500, 
new best solutions were discovered.
However the successive RPD values (see Section~\ref{sec:experimentalRes} Tables~\ref{tab:res1} and \ref{tab:res2}) show that \IGl~ still improves the solution when the original problem is considered ($\sigma=\infty$). 
This suggests \IGl~may be improved.
For instances of size 500, we noticed that the initial solutions used to identify the super-jobs are very diversified (the size of the problem was barely reduced by two with the lower level of confidence $\sigma=60$) because the original IG is not efficient for this size. Undoubtedly, this has an effect on the performance of the whole execution of \IGl.  
Iterating \IGl~from the solutions obtained at the end of the 30 executions may improve the quality as well as the size of super-jobs identified  and so the performance of the algorithm.
%A reasonable idea is therefore to iteratively exploit the solutions reached by \IGl~ to identify better sub-sequences of jobs \ie new super-jobs, to use it in \IGl~ again to improve its performance.

%A single run of \IGl~ identifies the super-jobs from a pool of $R$ initial solutions. Therefore, $R$ solutions given by \IGl~ would be needed to iterate this process and so, $R*R$ solutions to iterate it again and so on.  $I$ iterations of \IGl~ would lead to generate $\sum^I_{k=1} R^k$ initial solutions in total. This direct approach of iterating \IGl~ is not an admissible proposition. If we look at the solutions given by 30 runs of \IGl~ in the experiments detailed in Section~\ref{sec:experiments}, we observe some similarities in the representation in terms of sub-sequences of jobs.  Therefore some (new) super-jobs could be identified, these solutions may constitute a new pool for a next iteration of \IGl. %
%\mek{jsuis pas tres sure de l'interet du paragraphe precedent ? est-ce qu'on irati pas direct sur la version iteree proposee ?!?}
The proposed iterative approach (\IIGl) given in Algorithm~\ref{algo:SJiterated} is based on this idea.

\begin{algorithm}
	\KwData{$\Sigma = \sigma_1, \sigma_2, \dots $: list of confidence levels in increasing order\;
	\hspace{1cm}$\mathcal{P}_0$: initial set of solutions\; 	
	\hspace{1cm}$\mathcal{P}_i$: set of solutions built at iteration $i$\; 
	\hspace{1cm}$\mathcal{P}_{tmp}$: temporary set of solutions used by \IGl\;	
	\hspace{1cm}$I$: number of iterations \; 
	\hspace{1cm}$R$: number of solutions built at each iteration $i$\; 
	\hspace{1cm}$\rho$: number of solutions used for learning\; 
	\hspace{1cm}$\pi$: solution\;
	\hspace{1cm}$\pi^*$: best solution.}

	$\pi^* = $best$(\mathcal{P}_0)$  \tcc*{Initialize the best solution with the best solutions of $\mathcal{P}_0$}
 	\For{$i$ in $1..I$}{
 		\tcp{INNER PROCEDURE}
 		$\mathcal{P}_i = \emptyset $  \tcc*{Initialize $\mathcal{P}_i$ as an empty set}
 		\For{$k$ in $1..R$} {
 			$\mathcal{P}_{tmp} = $ pick$(\rho, \mathcal{P}_{i-1})$  \tcc*{Pick $\rho$ solutions among $\mathcal{P}_{i-1}$ to be stored in $\mathcal{P}_{tmp}$ (i)}
 			$\pi = $ \IGl$(\mathcal{P}_{tmp}, \Sigma)$ \tcc*{(ii)}
 			$\mathcal{P}_i = \mathcal{P}_i \cup \pi$  \tcc*{Store $\pi$ in $\mathcal{P}_i$ (iii)}
 			$\pi^* = $best$(\pi, \pi^*)$ 	\tcc*{Memorize the best solution}
 		}
 	}
 	
 	\Return{$\pi^*$}
 	\caption{\IIGl: Iterative \IGl}
 	\label{algo:SJiterated}
\end{algorithm}

\begin{figure}
	\centering
	\includegraphics[scale=0.5]{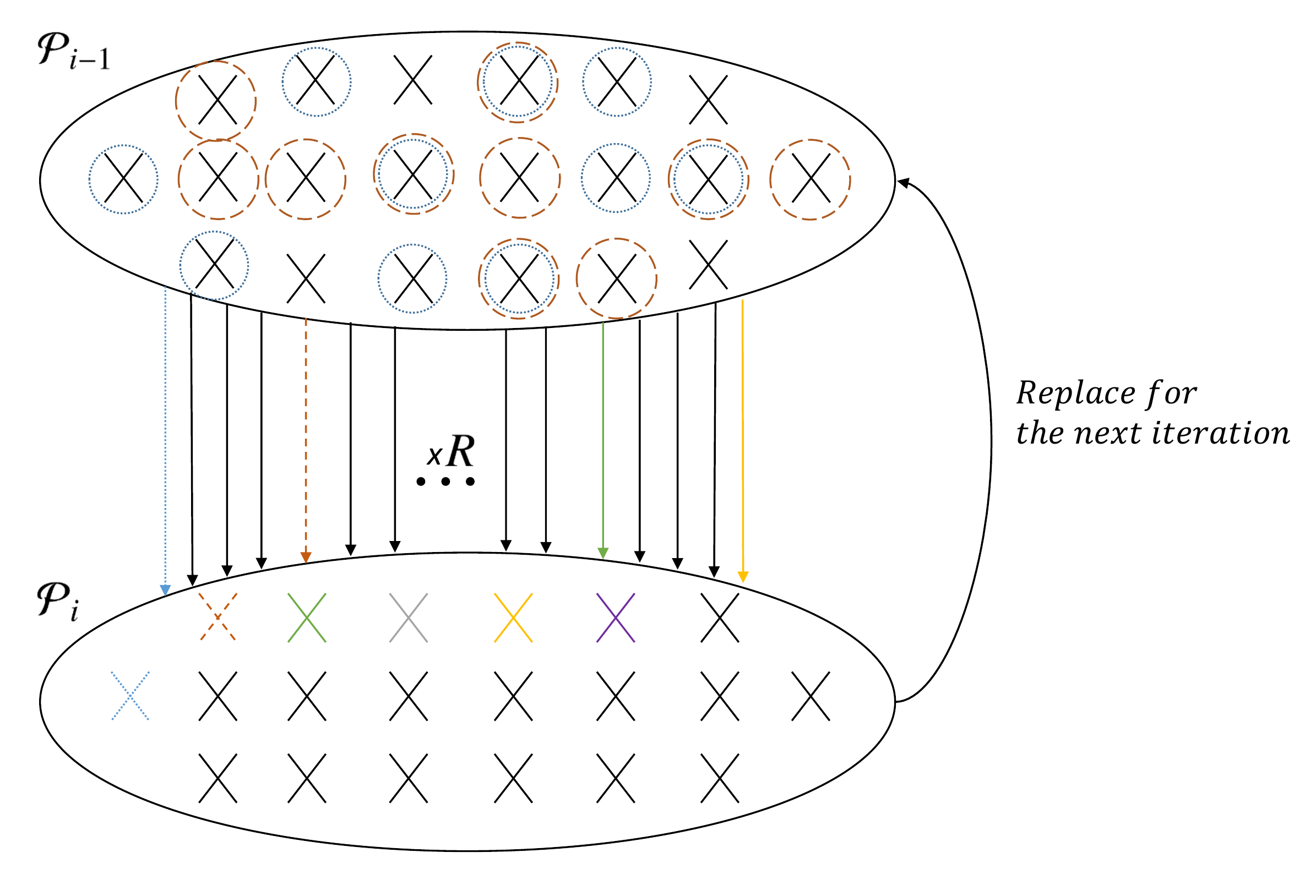} 
	\caption{Illustration of the inner procedure of \IIGl~ at iteration $i$. $\rho$ solutions (crosses encircled) are uniformly chosen at random in $\mathcal{P}_{i-1}$ and used to identify super-jobs. \IGl~ is run (down arrow) to give a new solution stored in $\mathcal{P}_i$. This process is repeated $R$ times. At the end of iteration $i$, $\mathcal{P}_i$ contains $R$ solutions and will replace $\mathcal{P}_{i-1}$ for the next iteration $i+1$.}
	\label{fig:SchemaIteratedSJ}
	
\end{figure}

\IIGl~ starts with a set $\mathcal{P}_0 $ of $R$ solutions, 
iterates $I$ times the inner procedure 
and returns the best solution $\pi^*$ found.
The inner procedure aims at building sets of solutions with better and better qualities
in order to identify super-jobs hopefully being those of the optimal solution.
At iteration $i\in I$, it starts with $\mathcal{P}_i$ as an empty set where $R$ new solutions will be iteratively added following these three steps:
(i) first, $\rho$ solutions are uniformly picked at random from the set $\mathcal{P}_{i-1}$ and stored in a temporary set $\mathcal{P}_{tmp}$ then 
(ii), \IGl~ is applied with $\mathcal{P}_{tmp}$ and $\Sigma$ to obtain a new solution $\pi$ that is finally 
(iii), stored in $\mathcal{P}_i$, the set of the current iteration $i$; 
if $\pi$ is better than the current $\pi^*$ then it replaces it. 
The parameter $\rho$ is used to select a subset of solutions and then to maintain the diversity in the constructed pool $\mathcal{P}_{tmp}$, otherwise the same super-jobs would be identified for each confidence level in phase (ii).
Figure~\ref{fig:SchemaIteratedSJ} gives an illustration of the inner procedure.

\subsection{Experimental Protocol \label{sec:IIGL_protocol}}

\IIGl~ presents its own parameters to settle in addition to those of \IGl.
The number of solutions $R$ of a pool $(\mathcal{P}_{i})_{i \geq 0}$ has been set to 20 that is statistically reasonable to evaluate the average performance of the approach; 
the number of solutions $\rho$ set to 10 like in the experiments of Section~\ref{sec:experiments} where 10 solutions were used to identify the super-jobs,
and the number of iterations $I$ set to 5, enough to converge and to prevent over-learning.
Iterated greedy (IG)~\citep{Ruiz_2007} is still the algorithm used in \IGl~ to improve solutions.
It is stopped when either a maximal time of $n_{sj} ^2$ ms (where $n_{sj}$ is the current number of super-jobs) 
or a maximal number of iterations without improvement of $25 * n$ is reached.
Note that this stopping criterion is shorter than the one set in the previous experiments (Section~\ref{sec:experiments}). 
Indeed, since the process is iterated, the end of the convergence is not mandatory for each execution of IG. 
Moreover, the use of \IGl~ requires the setting of the parameter $\Sigma$ to identify the jobs with different levels of confidence.
We use $\Sigma=\{60\%,70\%, 80\%, 90\%, \infty\}$ since the previous experiments showed better results for large size instances with this setting. % while results was equivalent between the two tested settings for the other ones.%small size instances.

\subsection{Experimental Results}

\begin{table}
	\resizebox{\textwidth}{!}{
	\begin{tabular}{lrr|lrr|lrr|lrr}
	\hline
Instance	&	Best	&	Gap	&	Instance	&	Best	&	Gap	&	Instance	&	Best	&	Gap	&	Instance	&	Best	&	Gap \\
\hline
Ta01	&	1,486	&	0	&	Ta31	&	\textbf{3,160}	&	-1	&	Ta61	&	\textbf{6,366}	&	-31	&	Ta91	&	\textbf{15,248}	&	-71\\
Ta02	&	1,528	&	0	&	Ta32	&	3,432	&	0	&	Ta62	&	\textbf{6,219}	&	-15	&	Ta92	&	\textbf{15,007}	&	-78\\
Ta03	&	1,460	&	0	&	Ta33	&	\textbf{3,210}	&	-1	&	Ta63	&	\textbf{6,108}	&	-13	&	Ta93	&	\textbf{15,276}	&	-100\\
Ta04	&	1,588	&	0	&	Ta34	&	\textbf{3,338}	&	-1	&	Ta64	&	\textbf{6,001}	&	-25	&	Ta94	&	\textbf{15,117}	&	-83\\
Ta05	&	1,449	&	0	&	Ta35	&	3,356	&	0	&	Ta65	&	\textbf{6,183}	&	-17	&	Ta95	&	\textbf{15,113}	&	-96\\
Ta06	&	1,481	&	0	&	Ta36	&	\textbf{3,346}	&	-1	&	Ta66	&	\textbf{6,058}	&	-16	&	Ta96	&	\textbf{14,997}	&	-112\\
Ta07	&	1,483	&	0	&	Ta37	&	3,231	&	0	&	Ta67	&	\textbf{6,224}	&	-23	&	Ta97	&	\textbf{15,300}	&	-95\\
Ta08	&	1,482	&	0	&	Ta38	&	3,235	&	0	&	Ta68	&	\textbf{6,115}	&	-15	&	Ta98	&	\textbf{15,162}	&	-75\\
Ta09	&	1,469	&	0	&	Ta39	&	\textbf{3,070}	&	-2	&	Ta69	&	\textbf{6,359}	&	-11	&	Ta99	&	\textbf{15,012}	&	-88\\
Ta10	&	1,377	&	0	&	Ta40	&	3,317	&	0	&	Ta70	&	\textbf{6,371}	&	-10	&	Ta100	&	\textbf{15,259}	&	-81\\
Ta11	&	2,044	&	0	&	Ta41	&	4,274	&	0	&	Ta71	&	\textbf{8,059}	&	-18	&	Ta101	&	\textbf{19,551} 	&	-130\\
Ta12	&	2,166	&	0	&	Ta42	&	4,177	&	0	&	Ta72	&	\textbf{7,859}	&	-21	&	Ta102	&	\textbf{19,980}	&	-116\\
Ta13	&	1,940	&	0	&	Ta43	&	4,099	&	0	&	Ta73	&	\textbf{8,017}	&	-11	&	Ta103	&	\textbf{19,791}	&	-122\\
Ta14	&	1,811	&	0	&	Ta44	&	4,399	&	0	&	Ta74	&	\textbf{8,330}	&	-18	&	Ta104	&	\textbf{19,775}	&	-153\\
Ta15	&	1,933	&	0	&	Ta45	&	4,322	&	0	&	Ta75	&	\textbf{7,939}	&	-19	&	Ta105	&	\textbf{19,732} 	&	-111  \\
Ta16	&	1,892	&	0	&	Ta46	&	4,289	&	0	&	Ta76	&	\textbf{7,773}	&	-28	&	Ta106	&	\textbf{19,852}	&	-90\\
Ta17	&	1,963	&	0	&	Ta47	&	4,420	&	0	&	Ta77	&	\textbf{7,851}	&	-15	&	Ta107	&	\textbf{19,967}	&	-145\\
Ta18	&	2,057	&	0	&	Ta48	&	4,318	&	0	&	Ta78	&	\textbf{7,881}	&	-32	&	Ta108	&	\textbf{19,900} 	&	-156\\
Ta19	&	1,973	&	0	&	Ta49	&	4,155	&	0	&	Ta79	&	\textbf{8,137}	&	-24	&	Ta109	&	\textbf{19,817} 	&	-101\\
Ta20	&	2,051	&	0	&	Ta50	&	4,283	&	0	&	Ta80	&	\textbf{8,095}	&	-19	&	Ta110	&	\textbf{19,794} 	&	-141\\
Ta21	&	2,973	&	0	&	Ta51	&	6,129	&	0	&	Ta81	&	\textbf{10,676}	&	-24	&	Ta111	&	\textbf{46,264}	&	-425\\
Ta22	&	2,852	&	0	&	Ta52	&	5,725	&	0	&	Ta82	&	\textbf{10,562}	&	-32	&	Ta112	&	\textbf{46,797}	&	-478\\
Ta23	&	3,013	&	0	&	Ta53	&	5,862	&	0	&	Ta83	&	\textbf{10,591}	&	-20	&	Ta113	&	\textbf{46,154}	&	-390\\
Ta24	&	3,001	&	0	&	Ta54	&	5,788	&	0	&	Ta84	&	\textbf{10,588}	&	-19	&	Ta114	&	\textbf{46,556}	&	-343\\
Ta25	&	3,003	&	0	&	Ta55	&	5,886	&	0	&	Ta85	&	\textbf{10,507}	&	-32	&	Ta115	&	\textbf{46,402}	&	-339\\
Ta26	&	2,998	&	0	&	Ta56	&	5,863	&	0	&	Ta86	&	\textbf{10,624}	&	-66	&	Ta116	&	\textbf{46,667}	&	-274\\
Ta27	&	3,052	&	0	&	Ta57	&	5,962	&	0	&	Ta87	&	\textbf{10,793}	&	-32	&	Ta117	&	\textbf{46,170}	&	-339\\
Ta28	&	2,839	&	0	&	Ta58	&	5,926	&	0	&	Ta88	&	\textbf{10,801}	&	-38	&	Ta118	&	\textbf{46,495}	&	-378\\
Ta29	&	3,009	&	0	&	Ta59	&	5,876	&	0	&	Ta89	&	\textbf{10,703}	&	-20	&	Ta119	&	\textbf{46,408}	&	-335\\
Ta30	&	2,979	&	0	&	Ta60	&	5,958	&	0	&	Ta90	&	\textbf{10,752}	&	-46	&	Ta120	&	\textbf{46,433}	&	-414\\

	\end{tabular}}
	
	\caption{Best known solutions of Taillard instances. A bold value indicates a new best solution was found out by our approach.}
    	\label{tab:bestKnown}
\end{table}

Table~\ref{tab:bestKnown} reports the best-known solutions of all Taillard instances 
and gives the gap value between the results obtained by our approach and the previous best-known solutions of the literature obtained by TMIIG~\citep{Ding_2015}.
A gap equal to 0 means \IIGl~ reaches the best-known solutions of TMIIG, 
and a strictly negative gap means it finds out a solution with a better quality \ie a new best-known solution.
This table shows that for the largest instances of size 100, 200 and 500 jobs, \IIGl~ improves the good results already obtained with \IGl~ and finds out new best-known solutions. 
The quality of the best solution has been improved up to 478 like for the instance Ta112 for example. 
Clearly, \IIGl~ is very efficient to solve uniform instances of the no-wait flowshop scheduling problem
like Taillard instances.
As far as we know, Table~\ref{tab:bestKnown} reports the values of the best-known solutions of the literature for the Taillard instances at this time.
The full description (schedule $+$ makespan) of the best solutions is given online\footnote{\url{blinded_webpage}}.

\begin{table}[h!]
\resizebox{\textwidth}{!}{              
 \begin{tabular}{l|c|cc|cc|cc|cc|cc||c}
    & Iter 0 & \multicolumn{2}{c|}{Iter 1} & \multicolumn{2}{c|}{Iter 2} &\multicolumn{2}{c|}{Iter 3}  & \multicolumn{2}{c|}{Iter 4}  &\multicolumn{2}{c||}{Iter 5}  &
Total\\					
Instances	&	RPD	&	time	(s)	&	RPD	&	time (s)	&	RPD	&	time	(s)	&	RPD	&	time	(s)	&	RPD	&	time	(s)&	RPD	&	time	(s)\\	
\texttt{20$\times$5}	&	0.000	&	14	&	0.000	&		14	&	0.000	&		14	&	0.000	&		14	&	0.000	&	14	&	0.000	&	71\\		
\texttt{20$\times$10}	&	0.000	&	14	&	0.000	&		14	&	0.000	&		14	&	0.000	&		14	&	0.000	&	14	&	0.000	&	71\\		
\texttt{20$\times$20}	&	0.015	&	15	&	0.001	&		14	&	0.000	&		14	&	0.000	&		14	&	0.000	&	14	&	0.000	&	71\\		
\texttt{50$\times$5}	&	0.276	&	129	&	0.073	&		112	&	0.029	&		103	&	0.020	&		101	&	0.013	&	100	&	0.009	&	545\\		
\texttt{50$\times$10}	&	0.128	&	124	&	0.059	&		107	&	0.037	&		101	&	0.020	&		97	&	0.019	&	98	&	0.019	&	527\\			
\texttt{50$\times$20}	&	0.120	&	111	&	0.034	&		100	&	0.016	&		97	&	0.010	&		96	&	0.010	&	94	&	0.010	&	497\\		
\texttt{100$\times$5}	&	1.015	&	742	&	0.317	&		370	&	0.162	&		253	&	0.134	&		236	&	0.107	&	229	&	0.102	&	1830\\		
\texttt{100$\times$10}	&	0.577	&	332	&	0.215	&		282	&	0.147	&		254	&	0.134	&		243	&	0.124	&	231	&	0.112	&	1341\\		
\texttt{100$\times$20}	&	0.587	&	324	&	0.222	&		265	&	0.159	&		244	&	0.137	&		226	&	0.127	&	206	&	0.124	&	1264\\		
\texttt{200$\times$10}	&	1.431	&	1835	&	0.441	&		1281	&	0.187	&		1084	&	0.102	&		1005	&	0.085	&	976	&	0.084	&	6181\\		
\texttt{200$\times$20}	&	1.180	&	1726	&	0.366	&		1269	&	0.206	&		1116	&	0.134	&		1048	&	0.105	&	908	&	0.082	&	6068\\		
\texttt{500$\times$20}	&	1.748	&	32906	&	0.598	&		20789	&	0.299	&		16668	&	0.157	&		14747	&	0.090	&	11728	&	0.055	&	96838\\		

 \end{tabular}}
        \caption{Analysis of the 5 iterations of \IIGl. Results are presented according to the 12 different sizes of the Taillard instances $N \times M$. 
         The RPD value of a run is computed from the best-known quality reported in Table\ref{tab:bestKnown}. Times are given in seconds. For each iteration, the reported values of RPD and time are the average values computed over 200 runs.}

        \label{tab:RPDIter}
\end{table}

\begin{figure}[h!]
	\subfloat[$50\times 05$]{
		\includegraphics[width=.30\textwidth]{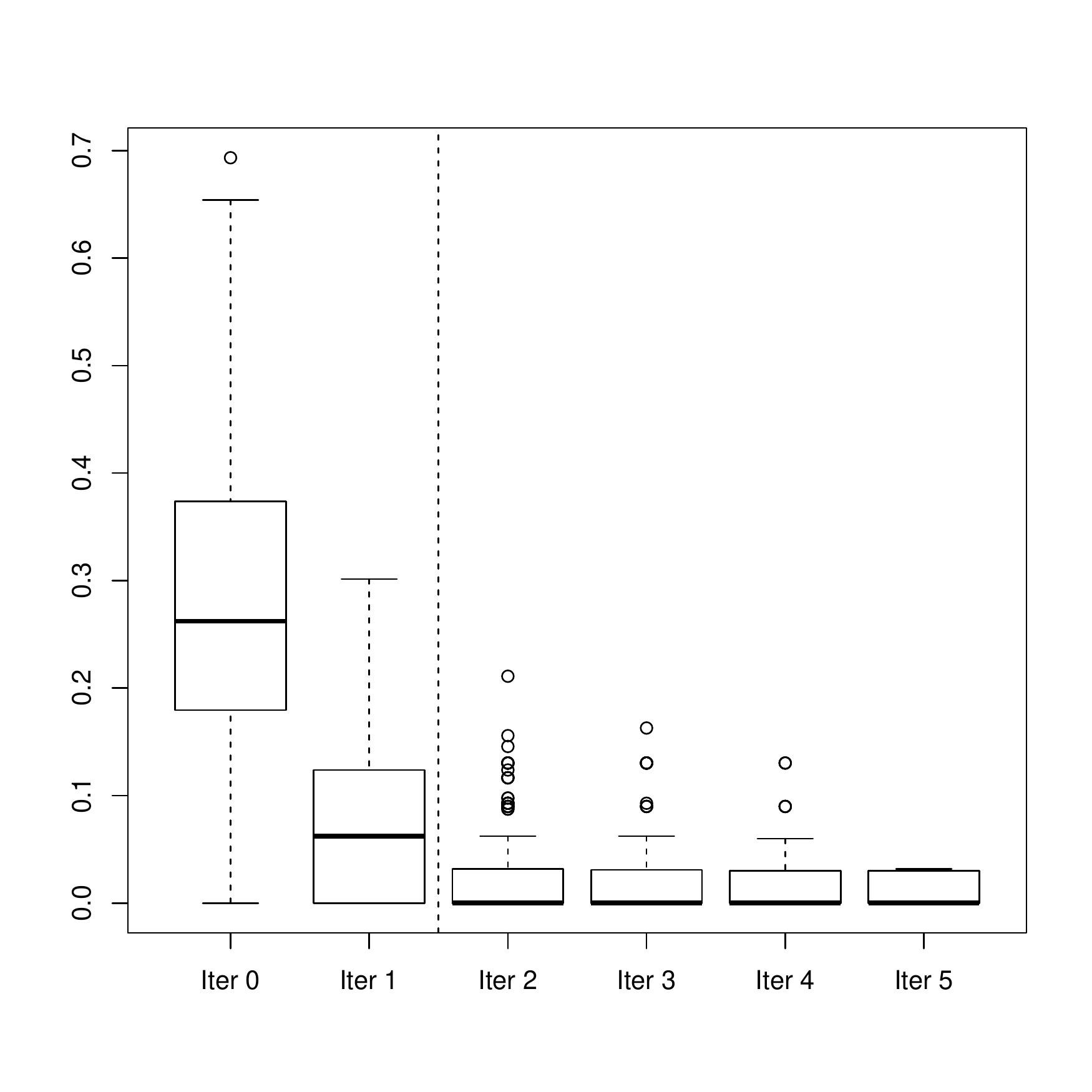}
	}
	\subfloat[$50\times 10$]{
		\includegraphics[width=.30\textwidth]{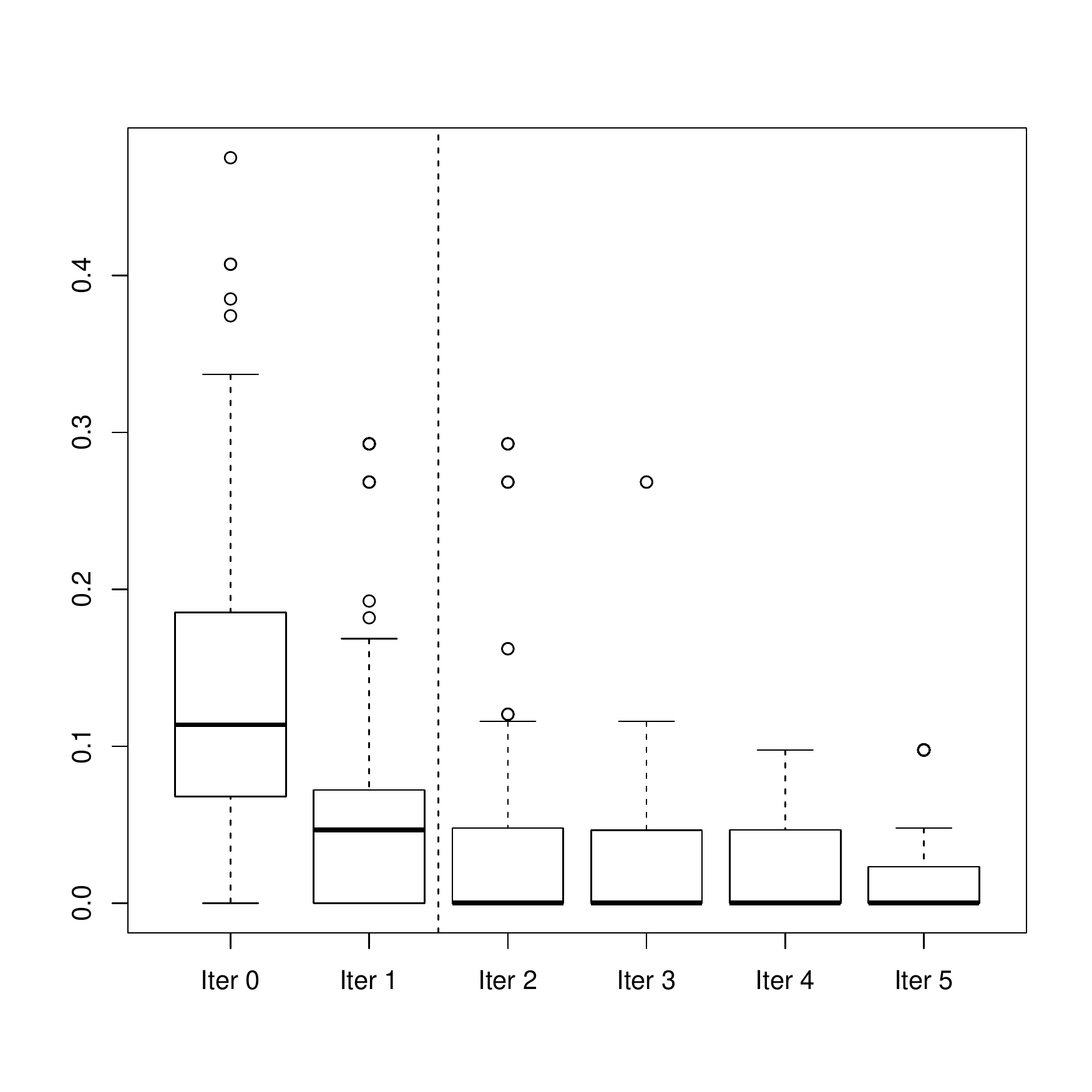}
	}
	\subfloat[$50\times 20$]{
		\includegraphics[width=.30\textwidth]{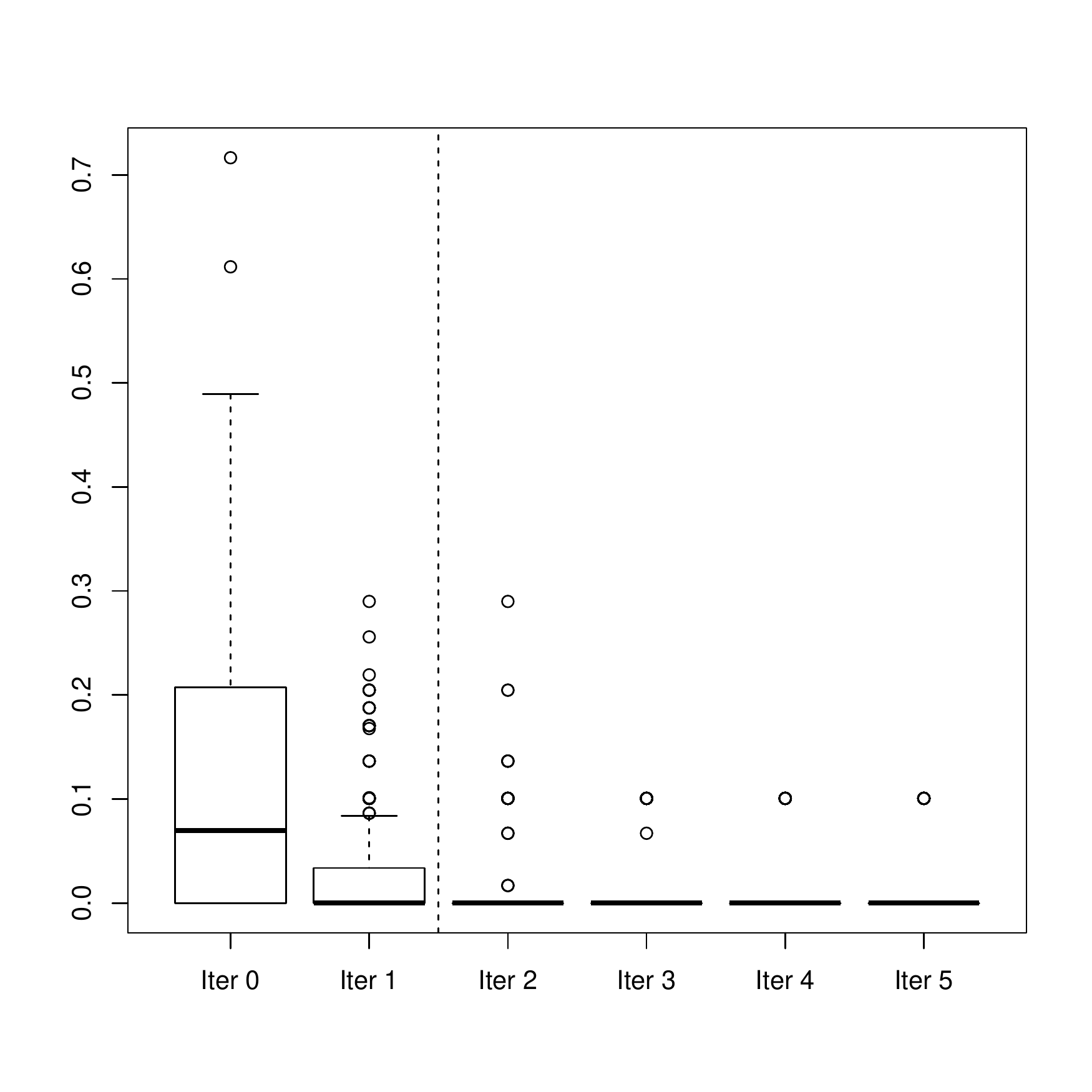}
	}
	
	\subfloat[$100\times 05$]{
		\includegraphics[width=.30\textwidth]{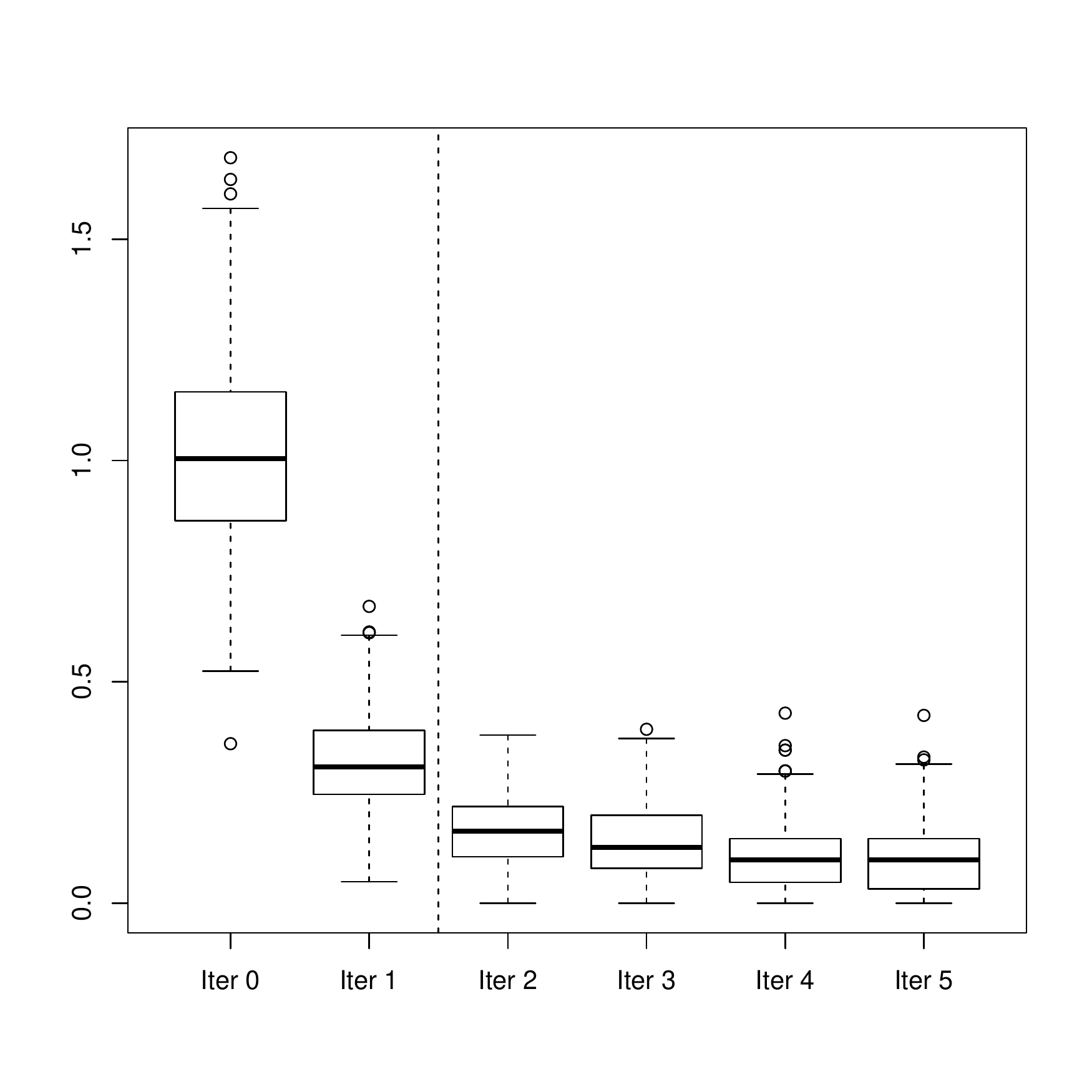}
	}
	\subfloat[$100\times 10$]{
		\includegraphics[width=.30\textwidth]{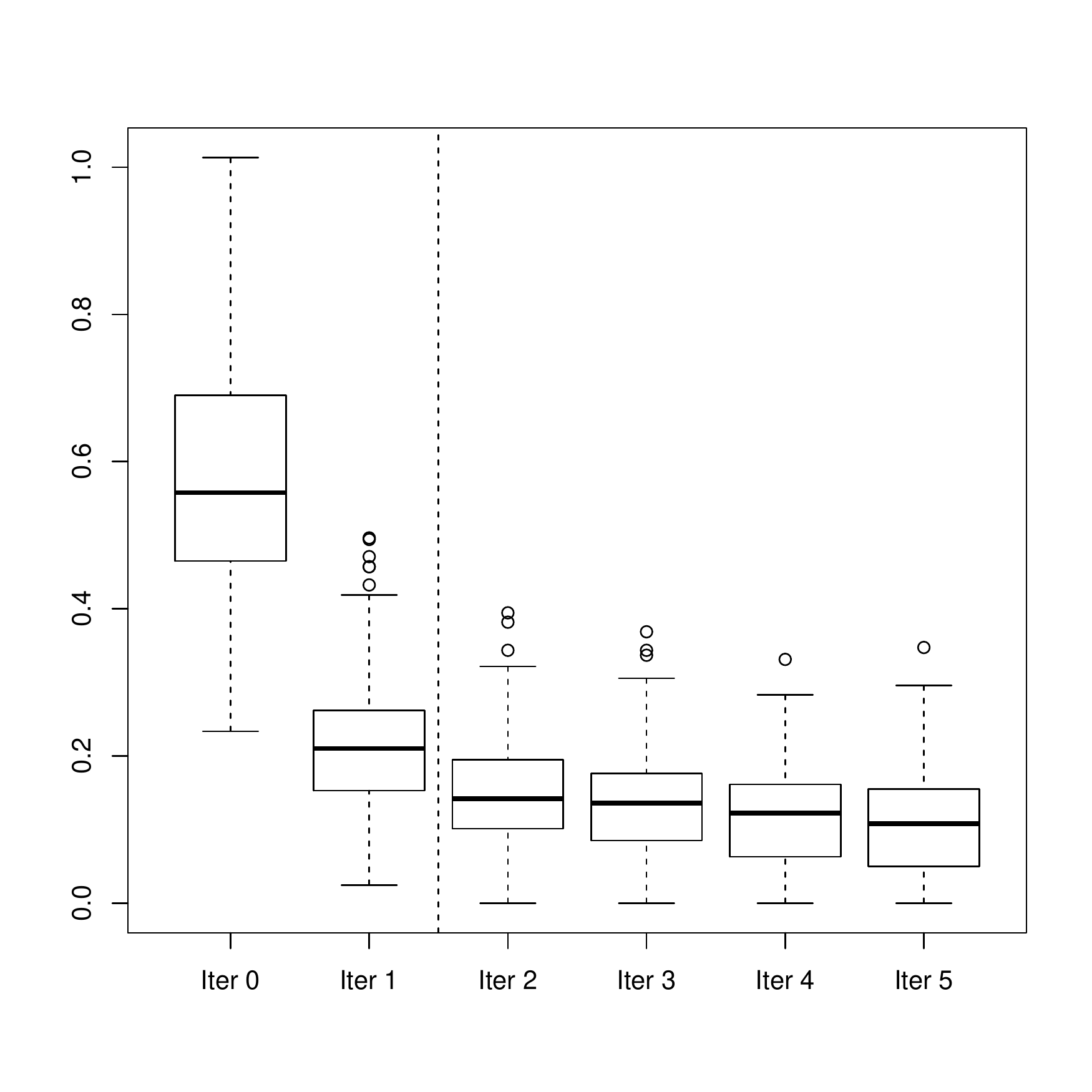}
	}
	\subfloat[$100\times 20$]{
		\includegraphics[width=.30\textwidth]{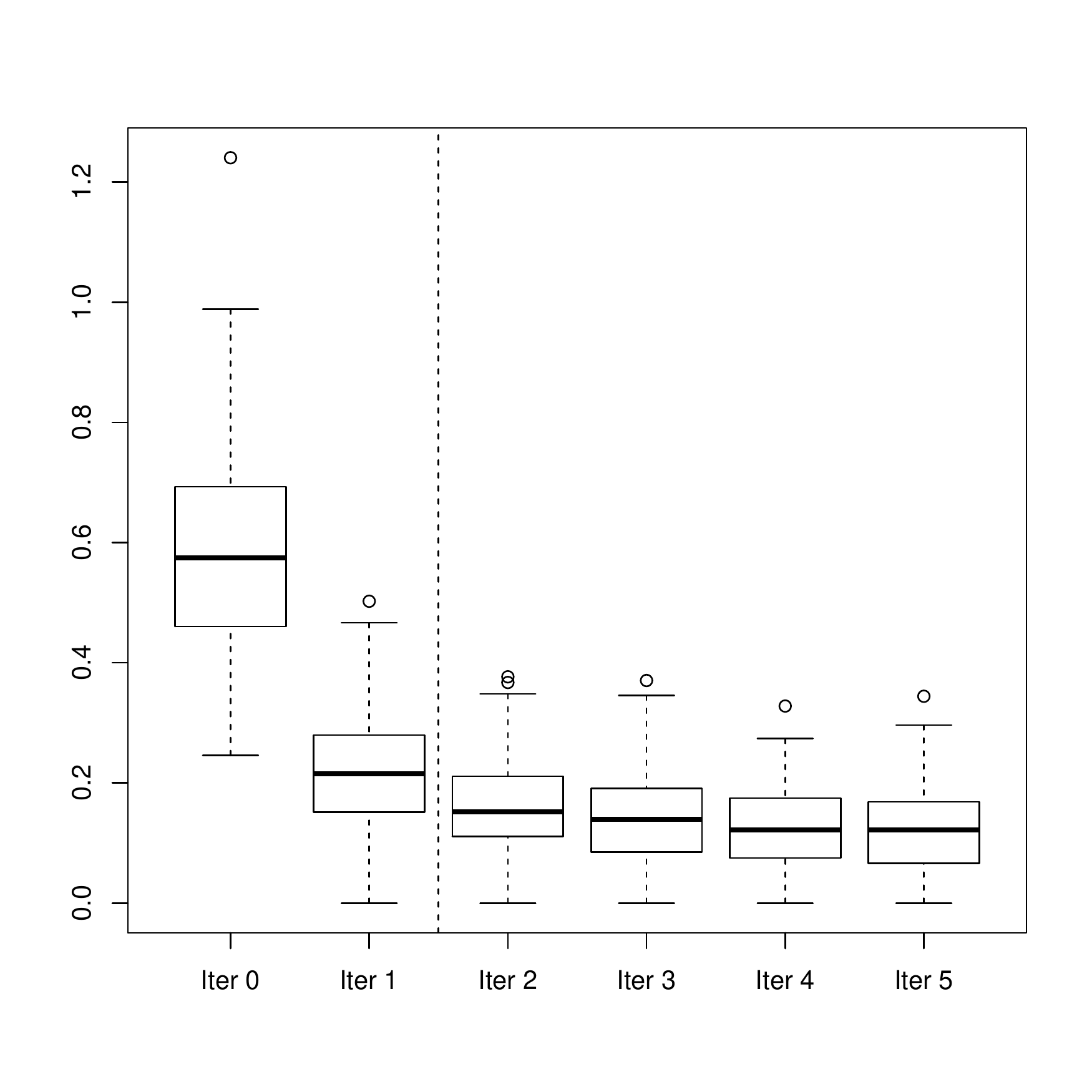}
	}
	
	\subfloat[$200\times 10$]{
		\includegraphics[width=.30\textwidth]{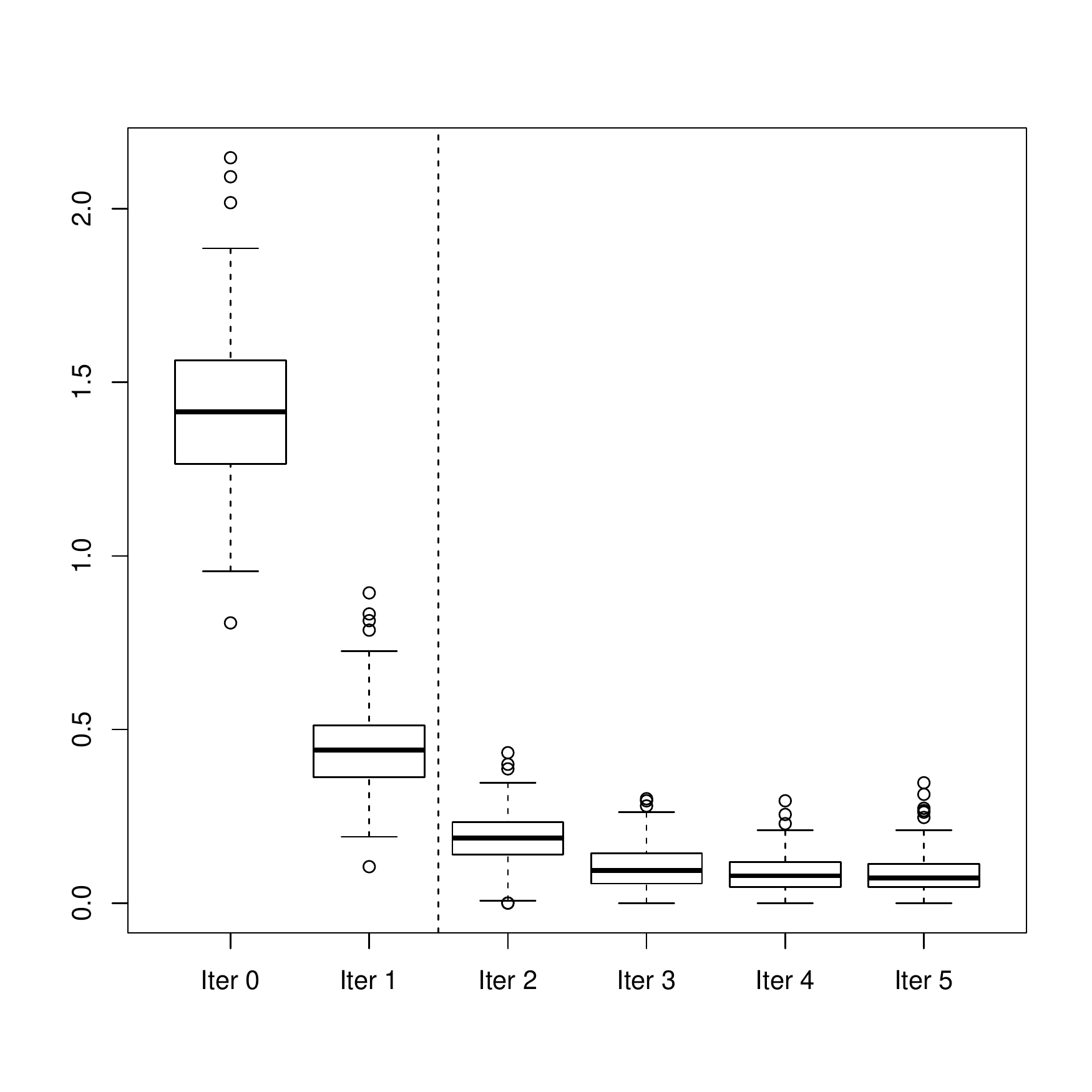}
	}
	\subfloat[$200\times 20$]{
		\includegraphics[width=.30\textwidth]{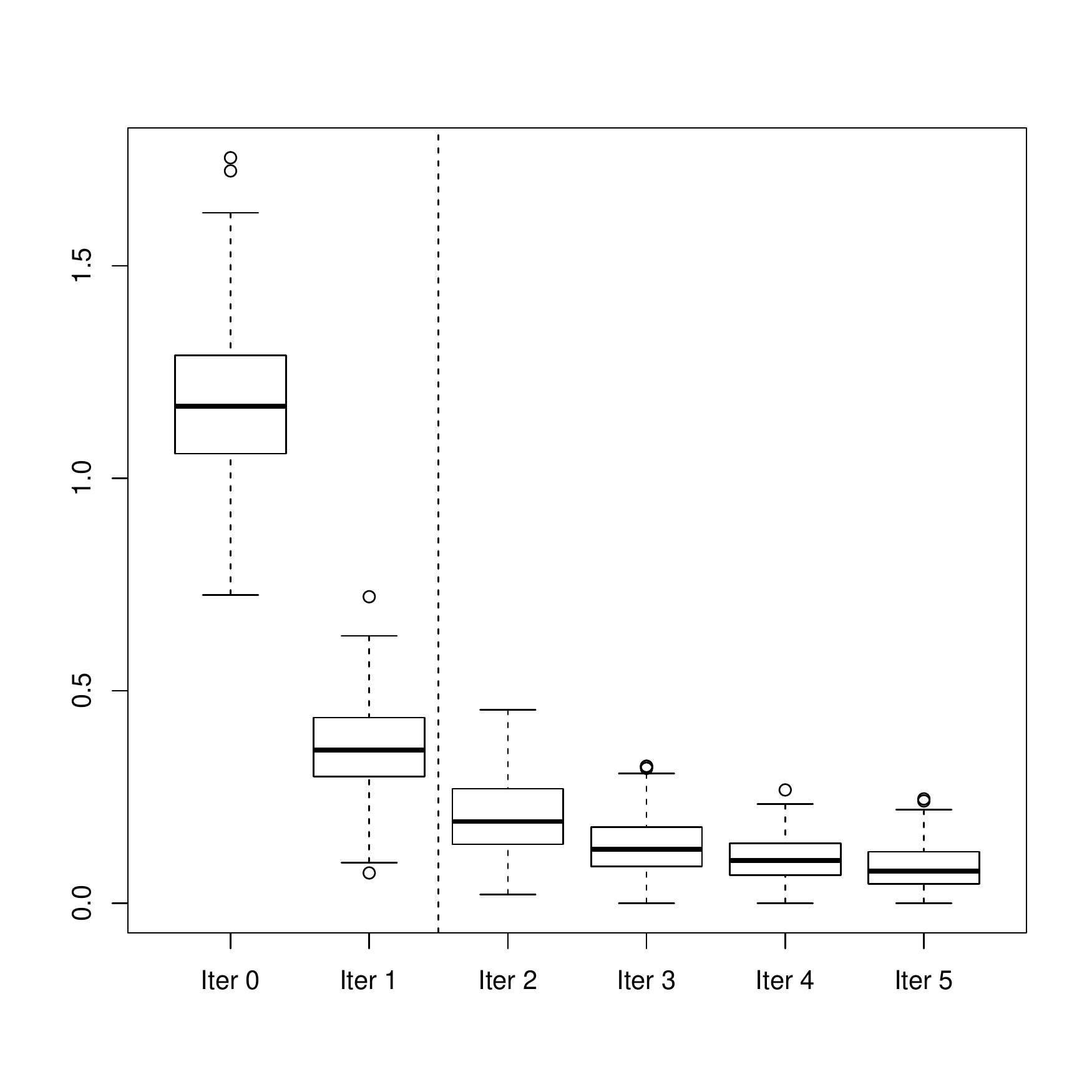}
	}
	\subfloat[$500\times 20$]{
		\includegraphics[width=.30\textwidth]{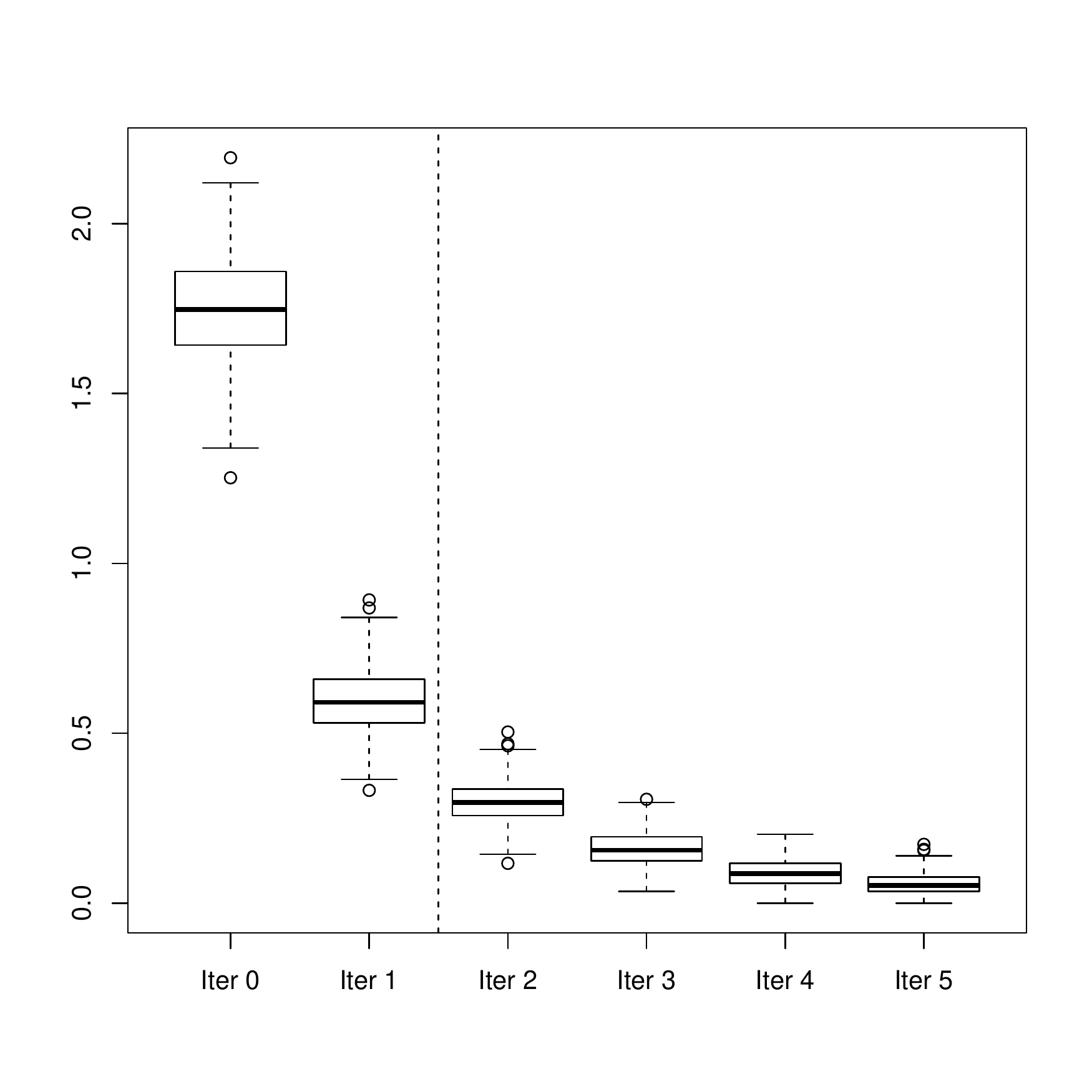}
	}
	\caption{Boxplots of the RPD values for 5 iterations of \IIGl.}
   \label{tab:boxplotsIter}
\end{figure}

In the following, we detail the results obtained by \IIGl~ 
and discuss the interest of using \IGl~ iteratively.
We present the results by grouping Taillard instances according to the 12 different sizes ($N \times M$) since the number of jobs $N$ and the number of machines $M$ impact the resolution of the problem.
In our experiment, one iteration of \IIGl~ provides 20 solutions.
All the solutions obtained after each iteration during a run of \IIGl~ are memorized in order to make an \emph{a posteriori} analysis to validate the interest of iterating the  approach of \IGl.
After running \IIGl~ on Taillard instances, the qualities of the solutions obtained after each iteration and the (new) best-known quality (values reported in Table~\ref{tab:bestKnown}) are compared to compute the RPD value (see Section~\ref{sec:protocol}).
Table~\ref{tab:RPDIter} gives the average RPD computed from the 200 values obtained (20 solutions per iteration, 10 instances by size) for each iteration of \IIGl.
A null RPD value means that the quality of the 200 solutions provided at the considered iteration is equal to the best-known quality for each of the 10 instances respectively.
A strictly positive RPD value means that at least one solution provided at the end of the iteration does not have the best-known quality.
As expected, the average RPD decreases with the successive iterations that shows the interest of exploiting the solution provided by an iteration to identify new and better super-jobs. This decrease is illustrated on Figure~\ref{tab:boxplotsIter} that shows the associated boxplots for the most difficult instances (20-jobs instances are optimally solved from the beginning).

A line separates each graphic: the left part corresponds to the first iteration (like one execution of \IGl) while the right part corresponds to the next iterations performed in \IIGl.
We observe that the larger the instance size, the larger the improvement of the median quality. 
%The boxplots show outlier solutions over the 200 for each iteration.
Between iteration 4 and iteration 5, for different instances (\emph{eg.} 100$\times$ 10, 100$\times$ 20, 200$\times$ 10), some qualities are even deteriorated. 
This may be explained by over-learning where the solutions in the sets $\mathcal{P}_{3}$ or $\mathcal{P}_{4}$ are too similar, and so the approach has difficulties to detect new super-jobs and get stuck in a particular region of the search space.    
Therefore the number of iterations of \IIGl~ has to be set carefully to avoid over-learning and a loss of time computing.

Table~\ref{tab:RPDIter} reports also, in seconds, the average execution time for each iteration and the average of the total time. Since the execution time depends on the number of (super-)jobs (see Section~\ref{sec:IIGL_protocol}), it increases a lot when the number of jobs increases. 
This value nearly reaches 96,838 seconds for the largest instances (size 500) \ie almost 28 hours.
Obviously, this time is not satisfactory practically. 
But, in these experiments, our goal is to simply improve the performance of our initial approach \IGl~ to find out new best solutions ; what was done. 
If we analyze more carefully the average times for each iteration, we observe a reduction of the time inversely proportional to the instance size, the larger this reduction the higher the number of jobs. We may explain this by the increase of the size of the super-jobs and the decrease of their numbers. Solutions are more and more similar within the set and so may share larger sub-sequences of jobs that gives a smaller and smaller number of jobs.

\clearpage
\newpage
%%%%%%%
\section{Conclusion \label{sec:conclusion}}

%%conclusion à partir de MIC
This paper presents \IGl~ a new approach based on the Iterated Greedy algorithm that found out 64 new best solutions of Taillard's instances for the No-Wait Permutation Flowshop Scheduling Problem (NWFSP).
The novelty of this approach is to modify the landscape during the search by first reducing the size of the initial problem and then, increasing it.
This process is allowed by the observed common structure of the best good quality solutions of the NWFSP that present common sub-sequences of consecutive jobs.
The identification of these sub-sequences has been discussed in the paper and has led to the definition of super-jobs with a confidence level.
Super-jobs are considered as a single job of the problem therefore, the confidence level indirectly determines the size of the problem to be solved.
The proposed approach consists in successively exploiting these super-jobs into the execution of Iterated Greedy.

In the experiments, we show that the number of super-jobs, identified with a middle level of confidence (here, 60\%) reduces by the half the size of the problem.
Therefore, the Iterated Greedy algorithm more easily discovers good regions of the search space.
The increase of the confidence level decomposes little by little the previous identified super-jobs hence, the search space is revealed gradually.

Besides, \IGl~ has been integrated in an iterative approach, called \IIGl, to take advantage of the solutions found after each execution to identify better and better super-jobs. Despite the high computational time of this method, the performance of \IGl~ has been successfully improved since the 64 new best solutions found out by \IGl have been overtaken. 
This work shows the benefit of using knowledge into optimization method such as meta-heuristics, and shows the impact of the quality of the knowledge used.

%%%
Future works will focus on how to adapt this approach for other permutation problems 
and so, how to identify other means to extract knowledge from the best solutions.
Moreover, the different increasing levels of confidence, needed to identify the super-jobs, will be analyzed to try to set the values in function of the instance.

% References

\bibliographystyle{unsrtnat}
\bibliography{biblio}

\begin{thebibliography}{23}
\providecommand{\natexlab}[1]{#1}
\providecommand{\url}[1]{\texttt{#1}}
\expandafter\ifx\csname urlstyle\endcsname\relax
  \providecommand{\doi}[1]{doi: #1}\else
  \providecommand{\doi}{doi: \begingroup \urlstyle{rm}\Url}\fi

\bibitem[R\"{o}ck(1984)]{Rock_1984}
Hans R\"{o}ck.
\newblock The three-machine no-wait flow shop is {NP}-complete.
\newblock \emph{Journal of the {ACM}}, 31\penalty0 (2):\penalty0 336--345, mar
  1984.
\newblock \doi{10.1145/62.65}.
\newblock URL \url{http://dx.doi.org/10.1145/62.65}.

\bibitem[Holger H.~Hoos(2005)]{Hoos_2004}
Thomas~St\"{u}tzle Holger H.~Hoos.
\newblock \emph{Stochastic Local Search: Foundations and Applications}.
\newblock Morgan Kaufmann Publishers Inc., 2005.
\newblock ISBN 1558608729.
\newblock URL
  \url{http://www.ebook.de/de/product/4330051/holger_h_hoos_thomas_stutzle_stochastic_local_search_foundations_and_applications.html}.

\bibitem[Taillard(1993)]{Taillard_1993}
E.~Taillard.
\newblock Benchmarks for basic scheduling problems.
\newblock \emph{European Journal of Operational Research}, 64\penalty0
  (2):\penalty0 278--285, jan 1993.
\newblock \doi{10.1016/0377-2217(93)90182-m}.
\newblock URL \url{http://dx.doi.org/10.1016/0377-2217(93)90182-m}.

\bibitem[Bertolissi(2000)]{Bertolissi_2000}
Edy Bertolissi.
\newblock Heuristic algorithm for scheduling in the no-wait flow-shop.
\newblock \emph{Journal of Materials Processing Technology}, 107\penalty0
  (1-3):\penalty0 459--465, nov 2000.
\newblock \doi{10.1016/s0924-0136(00)00720-2}.
\newblock URL \url{http://dx.doi.org/10.1016/s0924-0136(00)00720-2}.

\bibitem[Schiavinotto and St\"{u}tzle(2007)]{Schiavinotto_2007}
T.~Schiavinotto and T.~St\"{u}tzle.
\newblock A review of metrics on permutations for search landscape analysis.
\newblock \emph{Computers \& Operations Research}, 34:\penalty0 3143--3153,
  2007.

\bibitem[Kouvelis et~al.(2000)Kouvelis, Daniels, and
  Vairaktarakis]{Kouvelis_2000}
Panos Kouvelis, Richard~L. Daniels, and George Vairaktarakis.
\newblock Robust scheduling of a two-machine flow shop with uncertain
  processing times.
\newblock \emph{{IIE} Transactions}, 32\penalty0 (5):\penalty0 421--432, 2000.
\newblock \doi{10.1023/a:1007640726040}.

\bibitem[Pan et~al.(2007)Pan, Wang, Tasgetiren, and Zhao]{Pan_2007}
Quan-Ke Pan, Ling Wang, M.~Fatih Tasgetiren, and Bao-Hua Zhao.
\newblock A hybrid discrete particle swarm optimization algorithm for the
  no-wait flow shop scheduling problem with makespan criterion.
\newblock \emph{The International Journal of Advanced Manufacturing
  Technology}, 38\penalty0 (3-4):\penalty0 337--347, jul 2007.
\newblock \doi{10.1007/s00170-007-1099-4}.
\newblock URL \url{http://dx.doi.org/10.1007/s00170-007-1099-4}.

\bibitem[Nawaz et~al.(1983)Nawaz, Enscore, and Ham]{Nawaz_1983}
Muhammad Nawaz, E~Emory Enscore, and Inyong Ham.
\newblock A heuristic algorithm for the m-machine, n-job flow-shop sequencing
  problem.
\newblock \emph{Omega}, 11\penalty0 (1):\penalty0 91--95, jan 1983.
\newblock \doi{10.1016/0305-0483(83)90088-9}.
\newblock URL \url{http://dx.doi.org/10.1016/0305-0483(83)90088-9}.

\bibitem[Bianco et~al.(1999)Bianco, Dell'Olmo, and Giordani]{Bianco_1999}
Lucio Bianco, Paolo Dell'Olmo, and Stefano Giordani.
\newblock Flow shop no-wait scheduling with sequence dependent setup times and
  release dates.
\newblock \emph{{INFOR}: Information Systems and Operational Research},
  37\penalty0 (1):\penalty0 3--19, feb 1999.
\newblock \doi{10.1080/03155986.1999.11732365}.
\newblock URL \url{http://dx.doi.org/10.1080/03155986.1999.11732365}.

\bibitem[Gangadharan and Rajendran(1993)]{Gangadharan_1993}
Rajesh Gangadharan and Chandrasekharan Rajendran.
\newblock Heuristic algorithms for scheduling in the no-wait flowshop.
\newblock \emph{International Journal of Production Economics}, 32\penalty0
  (3):\penalty0 285--290, nov 1993.
\newblock \doi{10.1016/0925-5273(93)90042-j}.

\bibitem[Rajendran(1994)]{Rajendran_1994}
Chandrasekharan Rajendran.
\newblock A no-wait flowshop scheduling heuristic to minimize makespan.
\newblock \emph{Journal of the Operational Research Society}, 45\penalty0
  (4):\penalty0 472--478, apr 1994.
\newblock ISSN 1476-9360.
\newblock \doi{10.1057/jors.1994.65}.
\newblock URL \url{http://dx.doi.org/10.1057/jors.1994.65}.

\bibitem[Laha and Chakraborty(2008)]{Laha_2008}
Dipak Laha and Uday~K. Chakraborty.
\newblock A constructive heuristic for minimizing makespan in no-wait flow shop
  scheduling.
\newblock \emph{The International Journal of Advanced Manufacturing
  Technology}, 41\penalty0 (1-2):\penalty0 97--109, apr 2008.
\newblock \doi{10.1007/s00170-008-1454-0}.

\bibitem[Mousin et~al.(2017)Mousin, Kessaci, and Dhaenens]{Mousin_2017}
Lucien Mousin, Marie-El{\'{e}}onore Kessaci, and Clarisse Dhaenens.
\newblock A new constructive heuristic for the no-wait flowshop scheduling
  problem.
\newblock In \emph{Learning and Intelligent Optimization}, pages 196--209.
  Springer International Publishing, 2017.
\newblock \doi{10.1007/978-3-319-69404-7_14}.

\bibitem[Aldowaisan and Allahverdi(2003)]{Aldowaisan_2003}
Tariq Aldowaisan and Ali Allahverdi.
\newblock New heuristics for no-wait flowshops to minimize makespan.
\newblock \emph{Computers {\&} Operations Research}, 30\penalty0 (8):\penalty0
  1219--1231, jul 2003.
\newblock \doi{10.1016/s0305-0548(02)00068-0}.
\newblock URL \url{http://dx.doi.org/10.1016/s0305-0548(02)00068-0}.

\bibitem[Pan et~al.(2008)Pan, Tasgetiren, and Liang]{Pan_2008}
Quan-Ke Pan, M.~Fatih Tasgetiren, and Yun-Chia Liang.
\newblock A discrete particle swarm optimization algorithm for the no-wait
  flowshop scheduling problem.
\newblock \emph{Computers {\&} Operations Research}, 35\penalty0 (9):\penalty0
  2807--2839, sep 2008.
\newblock \doi{10.1016/j.cor.2006.12.030}.
\newblock URL \url{http://dx.doi.org/10.1016/j.cor.2006.12.030}.

\bibitem[Qian et~al.(2009)Qian, Wang, Hu, Huang, and Wang]{Qian_2009}
B.~Qian, L.~Wang, R.~Hu, D.X. Huang, and X.~Wang.
\newblock A {DE}-based approach to no-wait flow-shop scheduling.
\newblock \emph{Computers {\&} Industrial Engineering}, 57\penalty0
  (3):\penalty0 787--805, oct 2009.
\newblock \doi{10.1016/j.cie.2009.02.006}.
\newblock URL \url{http://dx.doi.org/10.1016/j.cie.2009.02.006}.

\bibitem[Grabowski and Pempera(2005)]{Grabowski_2005}
J{\'{o}}zef Grabowski and Jaros{\l}aw Pempera.
\newblock Some local search algorithms for no-wait flow-shop problem with
  makespan criterion.
\newblock \emph{Computers {\&} Operations Research}, 32\penalty0 (8):\penalty0
  2197--2212, aug 2005.
\newblock \doi{10.1016/j.cor.2004.02.009}.
\newblock URL \url{http://dx.doi.org/10.1016/j.cor.2004.02.009}.

\bibitem[Samarghandi and ElMekkawy(2012)]{Samarghandi_2012}
Hamed Samarghandi and Tarek~Y. ElMekkawy.
\newblock A meta-heuristic approach for solving the no-wait flow-shop problem.
\newblock \emph{International Journal of Production Research}, 50\penalty0
  (24):\penalty0 7313--7326, dec 2012.
\newblock \doi{10.1080/00207543.2011.648277}.
\newblock URL \url{http://dx.doi.org/10.1080/00207543.2011.648277}.

\bibitem[Jarboui et~al.(2010)Jarboui, Eddaly, and Siarry]{Jarboui_2010}
Bassem Jarboui, Mansour Eddaly, and Patrick Siarry.
\newblock A hybrid genetic algorithm for solving no-wait flowshop scheduling
  problems.
\newblock \emph{The International Journal of Advanced Manufacturing
  Technology}, 54\penalty0 (9-12):\penalty0 1129--1143, nov 2010.
\newblock \doi{10.1007/s00170-010-3009-4}.

\bibitem[Ding et~al.(2015)Ding, Song, Gupta, Zhang, Chiong, and Wu]{Ding_2015}
Jian-Ya Ding, Shiji Song, Jatinder~N.D. Gupta, Rui Zhang, Raymond Chiong, and
  Cheng Wu.
\newblock An improved iterated greedy algorithm with a tabu-based
  reconstruction strategy for the no-wait flowshop scheduling problem.
\newblock \emph{Applied Soft Computing}, 30:\penalty0 604--613, may 2015.
\newblock \doi{10.1016/j.asoc.2015.02.006}.
\newblock URL \url{http://dx.doi.org/10.1016/j.asoc.2015.02.006}.

\bibitem[Mladenovi{\'{c}} and Hansen(1997)]{Mladenovic_1997}
N.~Mladenovi{\'{c}} and P.~Hansen.
\newblock Variable neighborhood search.
\newblock \emph{Computers {\&} Operations Research}, 24\penalty0 (11):\penalty0
  1097--1100, nov 1997.
\newblock \doi{10.1016/s0305-0548(97)00031-2}.

\bibitem[Ruiz and St\"{u}tzle(2007)]{Ruiz_2007}
Rub{\'{e}}n Ruiz and Thomas St\"{u}tzle.
\newblock A simple and effective iterated greedy algorithm for the permutation
  flowshop scheduling problem.
\newblock \emph{European Journal of Operational Research}, 177\penalty0
  (3):\penalty0 2033--2049, mar 2007.
\newblock \doi{10.1016/j.ejor.2005.12.009}.
\newblock URL \url{http://dx.doi.org/10.1016/j.ejor.2005.12.009}.

\bibitem[Davendra et~al.(2013)Davendra, Zelinka, Bialic-Davendra, Senkerik, and
  Jasek]{Davendra_2013}
Donald Davendra, Ivan Zelinka, Magdalena Bialic-Davendra, Roman Senkerik, and
  Roman Jasek.
\newblock Discrete self-organising migrating algorithm for flow-shop scheduling
  with no-wait makespan.
\newblock \emph{Mathematical and Computer Modelling}, 57\penalty0
  (1-2):\penalty0 100--110, jan 2013.
\newblock \doi{10.1016/j.mcm.2011.05.029}.

\end{thebibliography}

% Appendix
\clearpage
\newpage
\vspace*{-10pt}
%%%%%%%
\appendix
\section{Value of best solutions found by the simple version algorithm \label{sec:annexe}}

Here are given the best solutions found by \IGl. Table \ref{tab:bestKnownConfig1} reports value obtained with the first configuration ($\Sigma = 60\%, 80\%, \infty$), and Table \ref{tab:bestKnownConfig2} with the second configuration ($\Sigma =60\%, 70\%, 80\%, 90\%, \infty$). A bold value indicates that a new best solution has been found. These tables do not report results for instances with 20 jobs, as for these small instances, optimal solutions are already found in the literature.

\begin{table}
	\resizebox{\textwidth}{!}{
	\begin{tabular}{lr|lr|lr|lr|lr|lr|lr}
	\multicolumn{2}{c|}{\texttt{050$\times$5}} & \multicolumn{2}{c|}{\texttt{100$\times$5}} & \multicolumn{2}{c|}{\texttt{100$\times$10}} & \multicolumn{2}{c|}{\texttt{100$\times$20}} & \multicolumn{2}{c|}{\texttt{200$\times$10}} & \multicolumn{2}{c|}{\texttt{200$\times$20}} & \multicolumn{2}{c}{\texttt{500$\times$20}}\\
Inst. &	Best & Inst. &	Best	&	Inst.	&	Best	&	Inst.	&	Best	&	Inst.	&	Best	&	Inst.	&	Best & Inst. & Best	\\
\hline																			
																									
Ta31	&	3,161	&	Ta61	&	\textbf{6,369}	&	Ta71	&	\textbf{8,062}	&	Ta81	&	\textbf{10,682}	&	Ta91	&	\textbf{15,279}	&	Ta101	&	\textbf{19,588} & Ta111	&	\textbf{46,454}	\\
Ta32	&	3,432	&	Ta62	&	\textbf{6,223}	&	Ta72	&	\textbf{7,866}	&	Ta82	&	\textbf{10,568}	&	Ta92	&	\textbf{15,058}	&	Ta102	&	\textbf{20,009} & Ta112	&	\textbf{46,934}	\\
Ta33	&	\textbf{3,210}	&	Ta63	&	\textbf{6,116}	&	Ta73	&	\textbf{8,022}	&	Ta83	&	\textbf{10,599}	&	Ta93	&	\textbf{15,282}	&	Ta103	&	\textbf{19,828} & Ta113	&	\textbf{46,329}	\\
Ta34	&	\textbf{3,338}	&	Ta64	&	\textbf{6,002}	&	Ta74	&	\textbf{8,334}	&	Ta84	&	\textbf{10,588}	&	Ta94	&	\textbf{15,148}	&	Ta104	&	\textbf{19,813} & Ta114	&	\textbf{46,743}	\\
Ta35	&	3,356	&	Ta65	&	\textbf{6,190}	&	Ta75	&	\textbf{7,939}	&	Ta85	&	\textbf{10,510}	&	Ta95	&	\textbf{15,136}	&	Ta105	&	\textbf{19,762} & Ta115	&	\textbf{46,563}	\\
Ta36	&	\textbf{3,346}	&	Ta66	&	\textbf{6,065}	&	Ta76	&	\textbf{7,780}	&	Ta86	&	\textbf{10,642}	&	Ta96	&	\textbf{15,032}	&	Ta106	&	\textbf{19,874} & Ta116	&	\textbf{46,816}	\\
Ta37	&	3,231	&	Ta67	&	\textbf{6,230}	&	Ta77	&	\textbf{7,851}	&	Ta87	&	\textbf{10,798}	&	Ta97	&	\textbf{15,334}	&	Ta107	&	\textbf{19,994} & Ta117	&	\textbf{46,365}	\\
Ta38	&	3,235	&	Ta68	&	\textbf{6,120}	&	Ta78	&	\textbf{7,886}	&	Ta88	&	\textbf{10,804}	&	Ta98	&	\textbf{15,193}	&	Ta108	&	\textbf{19,942} & Ta118	&	\textbf{46,694}	\\
Ta39	&	\textbf{3,070}	&	Ta69	&	\textbf{6,364}	&	Ta79	&	\textbf{8,143}	&	Ta89	&	\textbf{10,707}	&	Ta99	&	\textbf{15,041}	&	Ta109	&	\textbf{19,845} & Ta119	&	\textbf{46,589}	\\
Ta40	&	3,317	&	Ta70	&	\textbf{6,377}	&	Ta80	&	\textbf{8,096}	&	Ta90	&	\textbf{10,767}	&	Ta100	&	\textbf{15,282}	&	Ta110	&	\textbf{19,824} & Ta120	&	\textbf{46,629}	\\

	\end{tabular}}
	
	\caption{Solution fitness found with the first configuration ($60\% / 80\% / \infty$)}
    	\label{tab:bestKnownConfig1}
\end{table}

\begin{table}
	\resizebox{\textwidth}{!}{
	\begin{tabular}{lr|lr|lr|lr|lr|lr|lr}
	\multicolumn{2}{c|}{\texttt{050$\times$5}} & \multicolumn{2}{c|}{\texttt{100$\times$5}} & \multicolumn{2}{c|}{\texttt{100$\times$10}} & \multicolumn{2}{c|}{\texttt{100$\times$20}} & \multicolumn{2}{c|}{\texttt{200$\times$10}} & \multicolumn{2}{c|}{\texttt{200$\times$20}} & \multicolumn{2}{c}{\texttt{500$\times$20}}\\
Inst. &	Best & Inst. &	Best	&	Inst.	&	Best	&	Inst.	&	Best	&	Inst.	&	Best	&	Inst.	&	Best & Inst. & Best	\\
\hline																			
																									
Ta31	&	\textbf{3,160}	&	Ta61	&	\textbf{6,368}	&	Ta71	&	\textbf{8,063}	&	Ta81	&	\textbf{10,680}	&	Ta91	&	\textbf{15,280}	&	Ta101	&	\textbf{19,592} & Ta111	&	\textbf{46,466}	\\
Ta32	&	3,432	&	Ta62	&	\textbf{6,222}	&	Ta72	&	\textbf{7,859}	&	Ta82	&	\textbf{10,572}	&	Ta92	&	\textbf{15,052}	&	Ta102	&	\textbf{19,998} & Ta112	&	\textbf{46,949}	\\
Ta33	&	\textbf{3,210}	&	Ta63	&	\textbf{6,113}	&	Ta73	&	\textbf{8,020}	&	Ta83	&	\textbf{10,594}	&	Ta93	&	\textbf{15,286}	&	Ta103	&	\textbf{19,816} & Ta113	&	\textbf{46,326}	\\
Ta34	&	\textbf{3,338}	&	Ta64	&	\textbf{6,005}	&	Ta74	&	\textbf{8,334}	&	Ta84	&	\textbf{10,588}	&	Ta94	&	\textbf{15,143}	&	Ta104	&	\textbf{19,810} & Ta114	&	\textbf{46,724}	\\
Ta35	&	3,356	&	Ta65	&	\textbf{6,191}	&	Ta75	&	\textbf{7,939}	&	Ta85	&	\textbf{10,510}	&	Ta95	&	\textbf{15,128}	&	Ta105	&	\textbf{19,752} & Ta115	&	\textbf{46,560}	\\
Ta36	&	\textbf{3,346}	&	Ta66	&	\textbf{6,065}	&	Ta76	&	\textbf{7,779}	&	Ta86	&	\textbf{10,640}	&	Ta96	&	\textbf{15,028}	&	Ta106	&	\textbf{19,864} & Ta116	&	\textbf{46,810}	\\
Ta37	&	3,231	&	Ta67	&	\textbf{6,230}	&	Ta77	&	\textbf{7,855}	&	Ta87	&	\textbf{10,800}	&	Ta97	&	\textbf{15,334}	&	Ta107	&	\textbf{20,008} & Ta117	&	\textbf{46,320}	\\
Ta38	&	3,235	&	Ta68	&	\textbf{6,122}	&	Ta78	&	\textbf{7,885}	&	Ta88	&	\textbf{10,805}	&	Ta98	&	\textbf{15,185}	&	Ta108	&	\textbf{19,935} & Ta118	&	\textbf{46,666}	\\
Ta39	&	\textbf{3,070}	&	Ta69	&	\textbf{6,364}	&	Ta79	&	\textbf{8,143}	&	Ta89	&	\textbf{10,707}	&	Ta99	&	\textbf{15,046}	&	Ta109	&	\textbf{19,841} & Ta119	&	\textbf{46,553}	\\
Ta40	&	3,317	&	Ta70	&	\textbf{6,374}	&	Ta80	&	\textbf{8,096}	&	Ta90	&	\textbf{10,759}	&	Ta100	&	\textbf{15,283}	&	Ta110	&	\textbf{19,818} & Ta120	&	\textbf{46,591}	\\

	\end{tabular}}
	
	\caption{Solution fitness found with the second configuration ($60\% / 70\% / 80\% / 90\% / \infty$)}
    	\label{tab:bestKnownConfig2}
\end{table}

\end{document}